\documentclass[letterpaper]{article} 
\usepackage{aaai2026}  
\usepackage{times}  
\usepackage{helvet}  
\usepackage{courier}  
\usepackage[hyphens]{url}  
\usepackage{graphicx} 
\usepackage{natbib}  
\usepackage{caption} 
\usepackage{algorithm}
\usepackage{algorithmic}
\usepackage{longtable}
\usepackage{booktabs}
\usepackage{array}            
\usepackage{amsmath} 
\DeclareMathOperator*{\argmin}{arg\,min}

\usepackage{amsthm} 
\usepackage{amssymb} 
\usepackage{bbding}
\newtheorem{definition}{Definition}

\usepackage[most]{tcolorbox}
\usepackage{subcaption}
\usepackage{newfloat}
\usepackage{listings}
\DeclareCaptionStyle{ruled}{labelfont=normalfont,labelsep=colon,strut=off} 
\floatstyle{ruled}
\newfloat{listing}{tb}{lst}{}
\floatname{listing}{Listing}
\nocopyright

\title{LinkQA: Synthesizing Diverse QA \\from Multiple Seeds Strongly Linked by Knowledge Points}
\author{
    Xuemiao Zhang\textsuperscript{\rm 1,\rm 2}$^{\ast}$,  
    Can Ren\textsuperscript{\rm 1,\rm 2}\thanks{Equal contribution.},
    Chengying Tu\textsuperscript{\rm 1,\rm 2}$^{\ast}$, \\
    Rongxiang Weng\textsuperscript{\rm 2}$^{\dagger}$,
    Hongfei Yan\textsuperscript{\rm 1}\thanks{Corresponding author.}, 
    Jingang Wang\textsuperscript{\rm 2}, 
    Xunliang Cai\textsuperscript{\rm 2}
}
\affiliations{

    \textsuperscript{\rm 1} Peking University\quad
    \textsuperscript{\rm 2} Meituan\\
    \{zhangxuemiao, yanhf\}@pku.edu.cn\quad
    \{tuchengying, 2401210098\}@stu.pku.edu.cn\\
    wengrongxiang@gmail.com\quad
    \{wangjingang02, caixunliang\}@meituan.com
}

\usepackage{bibentry}

\begin{document}

\maketitle

\begin{abstract}
The advancement of large language models (LLMs) struggles with the scarcity of high-quality, diverse training data.
To address this limitation, we propose LinkSyn, a novel knowledge point (KP) graph-based synthesis framework that enables flexible control over discipline and difficulty distributions while balancing KP coverage and popularity. LinkSyn extracts KPs from question-answering (QA) seed data and constructs a KP graph to synthesize diverse QA data from multiple seeds strongly linked by KPs and sampled from graph walks. Specifically, LinkSyn incorporates
(1) a knowledge distribution value function to guide the adjustment of path sampling probability and balance KP coverage and popularity during graph walks; 
(2) diffusion-based synthesis via DeepSeek-R1 by leveraging multiple seeds with dense logical associations along each path;
and (3) high-difficulty QA enhancement within given disciplines by flexible difficulty adjustments.
By executing LinkSyn, we synthesize LinkQA, a diverse multi-disciplinary QA dataset with 50B tokens. Extensive experiments on Llama-3 8B demonstrate that continual pre-training with LinkQA yields an average improvement of \textbf{11.51\%} on MMLU and CMMLU, establishing new SOTA results. LinkQA consistently enhances performance across model size and initial FLOPs scales.\footnote{The core code and dataset will be made available at link.}

\end{abstract}

\section{Introduction}
As the scale of large language models (LLMs) escalates exponentially, the scarcity of high-quality training data has emerged as a critical bottleneck~\citep{muennighoff2025scalingdataconstrainedlanguagemodels,villalobos2024rundatalimitsllm}, particularly in multi-disciplinary domains~\citep{kandpal2023largelanguagemodelsstruggle}. Data synthesis has consequently gained prominence as a viable solution, offering scalable production of domain-specific knowledge representations~\citep{gunasekar2023textbooksneed,li2023textbooksneediiphi15,nadas2025syntheticdatagenerationusing}. Crucially, synthetic data in question–answer (QA) format has demonstrated significant efficacy in enhancing model performance on knowledge-intensive tasks by providing structured reasoning pathways and explicit knowledge representations~\citep{chen2024towards,wang2025octothinkermidtrainingincentivizesreinforcement,maini2024rephrasingwebrecipecompute}.

Despite these advances, current synthesis methods that depend on seed corpora face significant limitations. 
First, single-seed synthesis using trained models often experiences limited diversity due to inherent model biases~\citep{qin2025scalinglawssyntheticdata,su2024nemotron,akter2025mind,zhou2025megamath}. Second, entity-based methods~\citep{qin2025scalinglawssyntheticdata,Jiang2025SynthesizeonGraphKSA,yang2024syntheticcontinuedpretraining}, which extract sets of entities mentioned in documents and link documents through entity co-occurrence, aim to synthesize data from multiple connected documents. However, these methods exhibit limited knowledge integration, as individual entities rarely represent the document's core subject. Consequently, such connections lack semantic coherence, thus restricting cross-textual knowledge integration.
Additionally, current methods struggle to finely adjust the distributions of synthesized data in terms of difficulty, discipline, and knowledge popularity. This results in a low yield of valuable data and poor performance on benchmarks that require higher-order abilities, such as reasoning~\citep{hendrycks2021measuringmassivemultitasklanguage}.

\begin{figure}[t]
    \centering
    \includegraphics[width=0.45\textwidth]{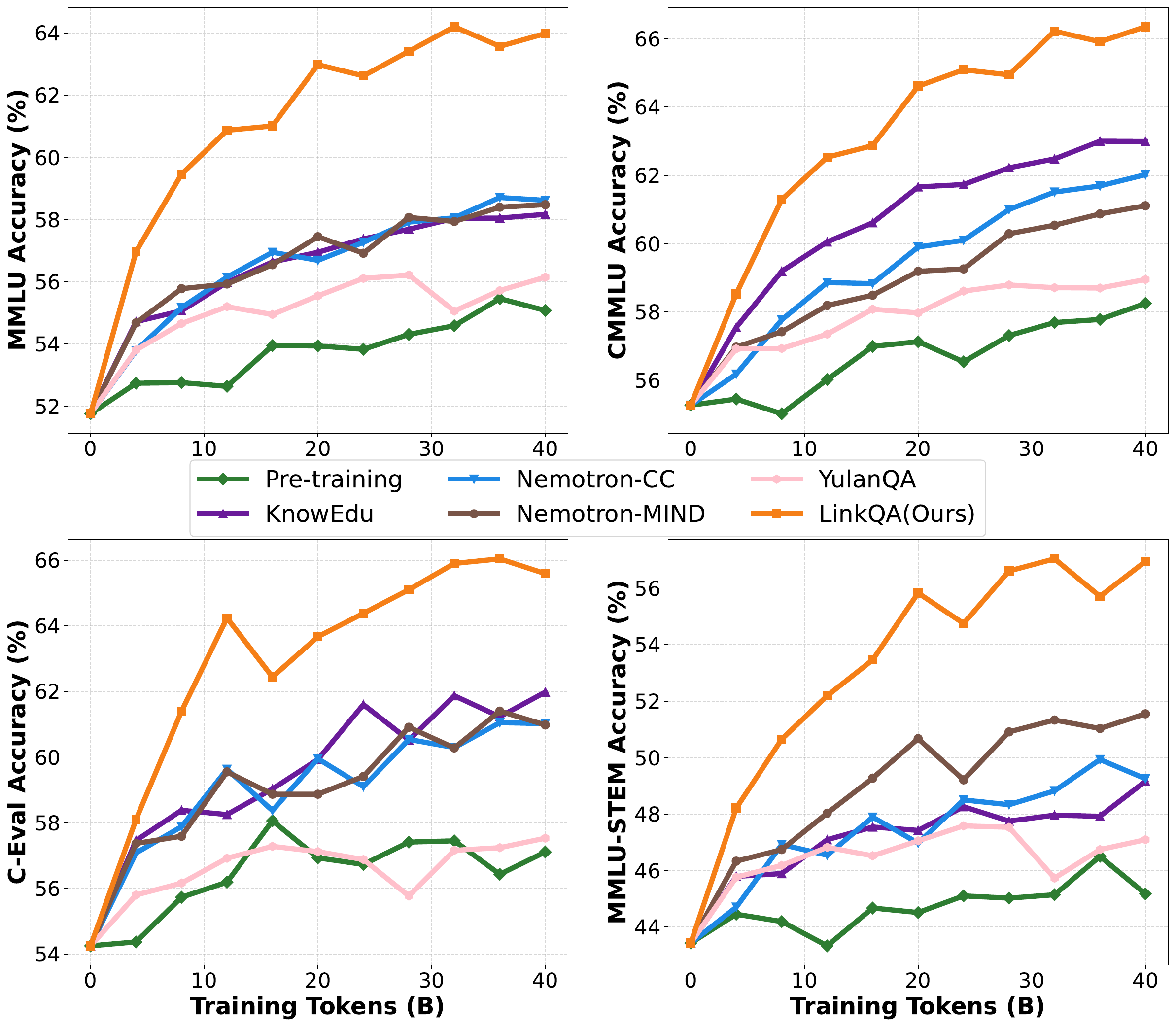}
    \caption{Comparison between LinkQA and baselines.}
    \label{fig:benchmark_comparison}
\end{figure}

Intuitively, unlike documents that mention numerous entities, a QA instance typically examines a few knowledge points (KPs), allowing each KP to serve as a strong representation of the QA itself. Thus, KP co-occurrence inherently provides more logical connections between QAs. Building on this insight, we propose constructing a KP graph from QA seeds to capture robust logical associations. We then introduce LinkSyn, a novel diversity-driven and theoretically rigorous synthesis framework that traverses this graph to generate multi-seed QAs with dense logical links.
Specifically, LinkSyn: (1) introduces knowledge distribution values to guide the adjustment of path sampling probability, balancing KP coverage and popularity during graph traversal; (2) synthesizes diverse or entirely novel QAs via DeepSeek-R1~\citep{deepseekai2025deepseekr1incentivizingreasoningcapability} in a diffusion-based manner by leveraging multiple seeds with dense logical associations along each path; and (3) enhances the concentration of high‑difficulty QAs within specified disciplines during synthesis by flexibly adjusting the difficulty levels and discipline proportions of the sampling seed instances along graph paths.

We conduct extensive experiments by continually pre-training Llama-3 8B~\citep{grattafiori2024llama3herdmodels} at the 2T-token checkpoint using LinkQA, a multi-disciplinary QA dataset synthesized by LinkSyn. As shown in Figure~\ref{fig:benchmark_comparison}, LinkQA achieves an average improvement of \textbf{11.51\%} over the pre‑training baseline on MMLU and CMMLU, attaining state‑of‑the‑art (SOTA) results. LinkQA also demonstrates scalable performance gains as model sizes expand from 1.7B to 16B and checkpoints progress from 2T-token to 10T-token. Our main contributions are summarized as follows:
\begin{itemize}
    \item We propose constructing a KP graph based on KPs extracted from seed QA data, and introduce a theoretically rigorous framework, LinkSyn, to synthesize diverse QAs from multiple seeds that are strongly linked by KPs.
    \item We dedicate substantial resources to executing LinkSyn to synthesize LinkQA, a diverse multi-disciplinary QA dataset with 50B tokens. LinkQA is controllable in terms of difficulty, discipline, and KP distributions,  fostering community research in scalable data synthesis and LLM advancement.
    \item Extensive experiments conducted on Llama-3 8B trained with 40B tokens demonstrate that our LinkQA improves by an average of \textbf{11.51\%} on MMLU and CMMLU and achieves SOTA average performance on 12 benchmarks.
\end{itemize}

\section{Method}
\begin{figure*}[ht]
    \centering
    \includegraphics[width=0.95\textwidth]{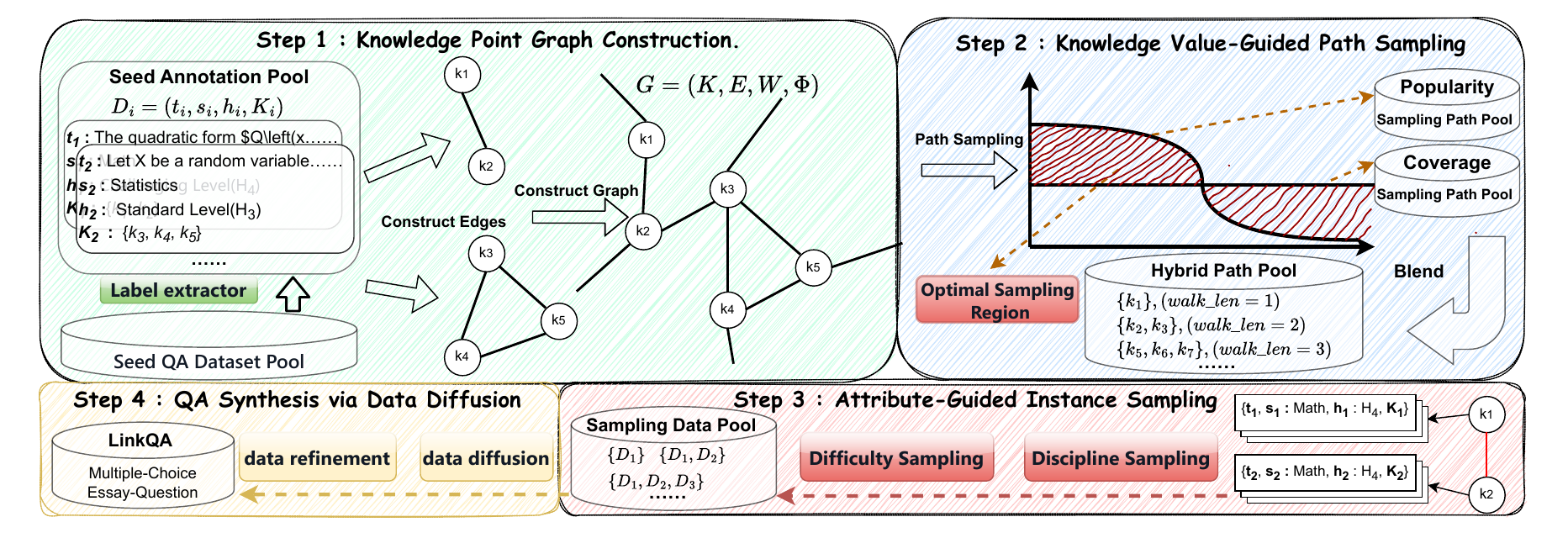}
    \caption{
    An overview of the LinkSyn pipeline. 
    \textbf{Step 1: Knowledge Point Graph Construction} — We construct the KP graph based on the co-occurrence of KPs in the seed data. 
    \textbf{Step 2: Knowledge Value-Guided Path Sampling} — Two sampling policies are utilized to balance KP coverage and popularity. 
    \textbf{Step 3: Attribute-Guided Instance Sampling} — Instances are sampled by controlling the distributions of difficulty and discipline.
    }
    \label{fig:pipeline}
\end{figure*}

\subsection{Overview}
LinkSyn is a framework designed to combine multiple seed QA instances to generate diverse samples that conform to expected distributions. As illustrated in Figure~\ref{fig:pipeline}, we first extract representative KPs from each QA instance and construct a KP graph where the edges denote strong logical relationships based on KP co-occurrence patterns. We then navigate this knowledge space using two complementary graph walking policies: popularity priority, which favors central KPs, and coverage priority, which explores diverse regions of the graph. For each KP along the generated paths, we sample seed instances according to specified difficulty levels and discipline distributions. Our diffusion-based approach then combines logically related instances to create novel QA pairs that preserve knowledge integrity.
\subsection{Knowledge Point Graph Construction}
\paragraph{Knowledge Point Extraction and Consolidation.} We define KPs as the fundamental units of content within a discipline, such as concepts, principles, theorems, or methods~\citep{duan2025enhancing,HAO202296}. For efficient and accurate annotation, we distill labeled data from DeepSeek-R1 and fine-tune the Qwen2.5-14B-Instruct model~\citep{qwen2025qwen25technicalreport} to serve as our extractor (see Appendix~\ref{sec:appendix_prompts_label}). We further consolidate the extracted KPs (see Appendix~\ref{app:dedup}), resulting in a final set of 10M\ high‑quality KPs.

Notably, 75.43\% of QA instances examine multiple KPs, indicating strong interrelations among KPs and motivating us to construct a KP graph with strong edge associations.
\paragraph{Knowledge Point Graph Construction.} The KP graph is built from the annotated QA dataset $A = \{D_i\}_{i = 1}^{m}$, where each item $D_i = (t_i, s_i, h_i, K_i)$ consists of the question text instance $t_i$, the discipline $s_i$, the difficulty $h_i$, and its associated KP set $K_i = \{k_{i1}, k_{i2}, \ldots, k_{in_i}\}$. We construct the KP graph $G = (K, E, W, \Phi)$ as follows: $K = \bigcup_{i=1}^m K_i$ is the set of unique KPs; $E$ is the set of undirected edges, where an edge $e_{k_p, k_q}$ exists if $k_p$ and $k_q$ co-occur in any instance:
$$
E = \{ e_{k_p, k_q} \mid k_p, k_q \in K,\, k_p \neq k_q,\, \exists D_l \in A: k_p, k_q \in K_l \}
$$
$W : E \rightarrow \mathbb{N}^+$ is the edge weight function, denoting the number of instances in which $k_p$ and $k_q$ co‑occur:
$$
W(e_{k_p, k_q}) = |\{ D_l \in A \mid k_p, k_q \in K_l \}|
$$
$\Phi : K \rightarrow 2^A$ maps each KP to the set of original data where it appears:
$$
\Phi(k) = \{ D_l \in A \mid k \in K_l \}
$$
An analysis of the constructed graph is presented in Appendix~\ref{sec:kp_graph_analysis}.
\subsection{Knowledge Value-Guided Path Sampling\label{sec:knowledge_value_optimization}}
\paragraph{Task Formulation.\label{para:knowledge_value_optimization}}
Based on the KP graph $G$, we design various random walk sampling policies $p$ to obtain a set $\Pi = \{\pi_i\}_{i=1}^M$ containing $M$ paths. Each path $\pi_i = \{k_{i1}, k_{i2}, \ldots, k_{il}\}$ consists of $l$ sequentially connected KPs. We deliberately set $l \in \{1, 2, 3\}$ to control the complexity of the generated questions.

\paragraph{Knowledge Distribution Value Optimization.} For simplicity, our analysis focuses on the distribution of KPs within $\Pi$. Let $N(k_i)$ denote the frequency with which the KP $k_i$ is sampled. Our goal is to devise a KP distribution that effectively balances two primary objectives:
\begin{itemize}
    \item \textbf{Coverage:} To enhance the representation of rare KPs~\citep{Huang_2025}, we define coverage as the expected count of distinct KPs sampled:
    \begin{equation}
    \begin{split}
    &\text{Coverage}(p) = \mathbb{E}\left[\sum_{k_i \in K} \mathbb{I}(N(k_i)>0)\right] \\
    &p^a = \arg\max_{p \in \Delta^{|K|}} \text{Coverage}(p) \\
    &p^a(k_i) = \frac{1}{|K|}, \forall k_i \in K
    \end{split}
    \label{eq:Coverage_eq}
    \end{equation}
    Here, $\mathbb{I}(\cdot)$ serves as the indicator function and $p^a$ represents the uniform distribution over all KPs, which maximizes coverage (see Appendix~\ref{app:coverage_uniform_proof}).
    
    \item \textbf{Popularity:} We propose aligning the sampled data with the real-world KP distribution~\citep{qin2025scalinglawssyntheticdata} by matching the empirical distribution:
    \begin{equation}
    p^b(k_i) = \frac{|\Phi(k_i)|}{\sum_{k_j \in K} |\Phi(k_j)|}.
    \label{eq:Popularity_eq}
    \end{equation}
\end{itemize}
To formalize the trade-off between coverage and popularity, we define the Knowledge Distribution Value as follows:
\begin{definition}[Knowledge Distribution Value]\mbox{}\\
$$
\mathrm{KV}(p) = \lambda D(p \,\|\, p^a) + (1 - \lambda)D(p \,\|\, p^b),
$$
where $p$ is the sampling probability distribution, $D(\cdot\,\|\,\cdot)$ is a divergence measure, $p^a$ is the uniform distribution over $K$, and $p^b$ is the empirical distribution.
\end{definition}
\paragraph{Sampling Policy.}
We aim to find an optimal sampling policy $p^*$ that balances coverage and popularity. Formally, 
\begin{equation}
p^* = \arg\min_{p} \mathrm{KV}(p).
\end{equation}
When the divergence $D(\cdot\|\cdot)$ is chosen as either the squared Euclidean distance or the (reverse) Kullback-Leibler (KL) divergence, the optimal solution $p^*$ takes the form $p^* = \alpha p^a + (1-\alpha) p^b$ (proof available in Appendix~\ref{app:kv_optimal_proof}).

We generalize the KP-based sampling probability to the random walk framework, leading to:

\begin{itemize}
    \item \textbf{Coverage Sampling Policy ($p^{a}$):} The starting node $k_1$ is selected uniformly at random (Eq.~\ref{eq:Coverage_eq}). The $(t+1)$-th node $k_{t+1}$ is selected uniformly among its neighbors:
        \begin{equation}
        \begin{aligned}
            p^{a}(k_{t+1} \mid k_t) &= \frac{1}{|\mathcal{N}(k_t)|}, \quad \forall k_{t+1} \in \mathcal{N}(k_t)
        \end{aligned}
        \end{equation}
        where $\mathcal{N}(k_t)$ denotes the set of neighbors of $k_t$ in $G$.

    \item \textbf{Popularity Sampling Policy ($p^{b}$):} The starting node $k_1$ is selected by its empirical frequency in the original dataset (Eq.~\ref{eq:Popularity_eq}). The $(t+1)$-th node $k_{t+1}$ is selected by the empirical co-occurrence frequency:
        \begin{equation}
        \begin{aligned}
            p^{b}(k_{t+1} \mid k_t) &= \frac{W(e_{k_t, k_{t+1}})}{\sum_{k' \in \mathcal{N}(k_t)} W(e_{k_t, k'})}
        \end{aligned}
        \end{equation}
\end{itemize}
\paragraph{Hybrid Sampling Policy.} 
We independently sample paths using $p^a$ and $p^b$, then blend the two sets of paths:
\begin{equation}
\Pi_{\text{hybrid}} = \alpha \cdot \Pi_{p^a} + (1-\alpha) \cdot \Pi_{p^b}, \quad \alpha \in [0,1]
\label{eq:hybrid_equation}
\end{equation}
By tuning $\alpha$, we can explore the trade-off between coverage and popularity, potentially improving model performance.
\subsection{Attribute-Guided Instance Sampling}
\paragraph{Task Formulation.}
Given the set of sampled KP paths $\Pi$, we construct seed datasets $\mathcal{S} = \{ S_\pi \mid \pi \in \Pi \}$, where
$$
S_\pi = \{t_{i} \mid D_{i} = (t_{i}, s_{i}, h_{i}, K_{i}) \in \Phi(k_i), k_i \in \pi\}
$$
Each text $t_{i}$ is sampled from $\Phi(k_i)$ and is required to satisfy the target attribute distributions.

\paragraph{Distribution Constraints.} We specify two attribute distributions: discipline and difficulty. Regarding difficulty, the original corpus contains a limited proportion of high-difficulty data (see Appendix~\ref{sec:Seed_Data_Distribution}). To rectify this imbalance and improve performance on challenging problems~\citep{tong2024dartmathdifficultyawarerejectiontuning}, we sample difficulty levels with the following probabilities: 10\% for H1, 15\% for H2, and 25\% for each of H3, H4, and H5, representing a gradient from easiest to hardest. For discipline, we focus on mathematics to create the LinkQA$_\text{Math}$ subset, which facilitates targeted evaluation of mathematical reasoning~\citep{huang2024subjectdrivescalinggenerativedata}.

\paragraph{Difficulty and Discipline Annotation.}
We categorize disciplines into 62 first-level categories and calibrate difficulty levels across five scales. We fine-tune Qwen2.5-7B-Instruct distilled from DeepSeek-R1 for large-scale automated labeling (detailed in Appendix~\ref{sec:appendix_prompts_label}).

\paragraph{Instance Selection.} For each node $k_i$ in the sampled path, we select a supporting instance $t^*$ from the set $\Phi(k_i)$ with difficulty $h_D$ closest to the target difficulty $h$, prioritizing those with matching discipline $s$:
\begin{equation}
t^* = \argmin_{\substack{t\ \text{in}\ D \in \Phi(k_i),  s_D = s~\text{if possible}}} |h_D - h|
\label{eq:instance-selection}
\end{equation}
where $h$ and $s$ denote the target difficulty and discipline.

\begin{algorithm}
\caption{KP Path Sampling and Seed Instance Selection}
\label{alg:kp-sampling}
\textbf{Input:} KP graph $G = (K, E, W, \Phi)$; sampling policies $p^a$ and $p^b$; mixing parameter $\alpha$; difficulty distribution $\rho_h$; discipline distribution $\rho_s$; path length $l$; path sample number $M$ \\
\textbf{Output:} Sampled instances $\mathcal{S}$
\begin{algorithmic}
\STATE \textbf{Function} PS$(G, p, l, M)$: // Path Sampling
\STATE \quad Initialize $\Pi \gets \emptyset$ \quad // Set of sampled paths
\STATE \quad \textbf{while} $|\Pi| < M$ \textbf{do}
\STATE \quad\quad $k_1 \sim p(k_1)$ on $K$; $\pi \gets [k_1]$ 
\STATE \quad\quad \textbf{for} $t = 1$ to $l-1$ \textbf{do} \textbf{if} $N(k_t) \neq \emptyset$ \textbf{then} sample $k_{t+1}$ from $N(k_t)$ with $p(k_t, k_{t+1})$; append $k_{t+1}$ to $\pi$
\STATE \quad\quad \textbf{if} $\pi \notin \Pi$ \textbf{then} add $\pi$ to $\Pi$
\STATE \quad \textbf{return} $\Pi$
\STATE $\Pi^a \gets \text{PS}(G, p^a, l, M)$; $\Pi^b \gets \text{PS}(G, p^b, l, M)$
\STATE $\Pi_{\text{hybrid}} = \alpha \cdot \Pi_{p^a} + (1-\alpha) \cdot \Pi_{p^b}, \quad \alpha \in [0,1]$
\STATE Initialize $\mathcal{S} \gets \emptyset$ \quad // Set of sampled instances
\STATE \textbf{for} each path $\pi$ in $\Pi_{\text{hybrid}}$
    \STATE \quad Initialize $\mathcal{S}_\pi \gets \emptyset$; Sample $h \sim \rho_h$, $s \sim \rho_s$
    \STATE \quad \textbf{for} each node $k_t$ in $\pi$
    \STATE \quad \quad Sample $t^*$ according to Eq.~\eqref{eq:instance-selection} with $k_t$, $h$, and $s$ from instances not in $\mathcal{S}_\pi$
    \STATE \quad \quad Add $t^*$ to $\mathcal{S}_\pi$
    \STATE \quad \textbf{if} $\mathcal{S}_\pi \notin \mathcal{S}$ \textbf{then} add $\mathcal{S}_\pi$ to $\mathcal{S}$
\RETURN $\mathcal{S}$
\end{algorithmic}
\end{algorithm}

\subsection{QA Synthesis via Data Diffusion}
The algorithm for obtaining related seed sets via path and instance sampling is described in Algorithm~\ref{alg:kp-sampling}. We sample 20M seed groups for each combination of random walk length $l \in \{1, 2, 3\}$ and sampling policy, and additionally perform mathematics‑constrained sampling for the LinkQA$_\text{Math}$ subset, and then blend them in equal proportions ($\alpha = 0.5$). Utilizing these seed data, we employ DeepSeek-R1 for QA synthesis and DeepSeek-V3~\citep{deepseekai2025deepseekv3technicalreport} for answer refinement (details in Appendix~\ref{sec:appendix_prompts_synthesis}). Subsequently, we perform comprehensive data cleaning, including the mitigation of benchmark contamination through embedding‑based similarity and 10‑gram matching filters~\citep{shao2024deepseekmathpushinglimitsmathematical}, as well as low‑quality data filtering (see Appendix~\ref{sec:appendix_qualityreview}). This rigorous pipeline ultimately yields the high-quality 50B-token LinkQA dataset. Finally, we blend LinkQA with high-quality corpora, KnowEdu, to construct the training dataset. KnowEdu is curated from pre-training corpora, where the QuRater~\citep{wettig2024qurating} quantifies knowledge density and the educational classifier from FineWeb-Edu~\citep{penedo2024fineweb} evaluates educational utility. Texts rated highly in both knowledge density and educational utility are retained to form KnowEdu.
\section{Experiments}

\subsection{Experimental Setup}

\paragraph{Training Details.} We use DeepSeek-R1 for data synthesis and DeepSeek-V3 for answer refinement on multiple H20 GPUs. The effectiveness of LinkQA is validated during continual pre-training using Llama-3 8B, which is pre-trained on 10T tokens. Our main experiments commence continual pre-training from the 2T-token checkpoint using a 1:1 mixture of QA and KnowEdu, with 40B tokens, implemented via the Megatron framework~\citep{shoeybi2020megatronlmtrainingmultibillionparameter} and optimized by the Adam algorithm. The training employs a linearly decaying learning rate schedule initialized at $1.9\times 10^{-4}$ and terminating at $1.9\times 10^{-5}$. Further scaling experiments systematically examine the model size scale by evaluating 1.7B and 16B architectures under identical 40B-token configurations, and initial FLOPs scale of 2T and 10T tokens for the 8B model. Details of the training setup are provided in Appendix~\ref{sec:appendix_train_details}.

\begin{table*}
\centering
\setlength{\tabcolsep}{1mm}
\small
\begin{tabular}{l|ccccccccccccc}
\toprule
\textbf{Dataset} & \textbf{MMLU} & \textbf{CMMLU} & \textbf{C-Eval} & \textbf{M-Pro} & \textbf{STEM} & \textbf{MATH} & \textbf{GSM8K} & \textbf{W.G.} & \textbf{H.S.} & \textbf{BBH} & \textbf{ARC-C} & \textbf{DROP} & \textbf{AVG.} \\
\midrule
Pre-training & 55.08	& 52.23	& 57.11	& 24.32	& 45.17	& 6.50	& 33.95	&51.50	& \textbf{43.00}	& 35.79	& 70.50	& 42.31	& 43.12 \\
FineWeb-Edu & 56.23	&58.88	&56.80	&25.46	&47.78	&2.50	&31.49	&53.50	&35.00	&34.38	&69.60	&39.44	&42.59 \\
KnowEdu & 58.17	&62.99	&61.98	&25.64	&49.16	&8.00	&32.56	&54.50	&36.00	&35.12	&71.50	&41.07	&44.72 \\
\midrule
Nemotron-CC	&58.62	&62.02	&59.02	&27.68	&49.25	&7.00	&29.33	&55.00	&41.50	&34.86	&73.00	&41.63	&44.91 \\
YulanQA	&56.15	&58.95	&57.53	&24.89	&47.09	&9.00	&30.48	&53.00	&\textbf{43.00}	&35.75	&73.00	&42.18	&44.25 \\
Nemotron-MIND	&58.48	&61.11	&60.98	&30.32	&51.55	&13.50	&47.42	&52.00	&40.50	&37.69	&73.00	&46.33	&47.74 \\
MegaMathQA	&55.98	&59.04	&58.91	&26.86	&48.59	&6.50	&44.65	&55.50	&41.50	&35.64	&68.50	&43.31	&45.42 \\
JiuZhang3.0	&56.55	&60.30	&59.52	&27.43	&48.68	&\textbf{23.00}	&\textbf{56.27}	&55.00	&36.50	&36.33	&71.50	&45.05	&48.01 \\
\midrule
LinkQA	&\textbf{63.98}	&\textbf{66.35}	&\textbf{65.59}	&\textbf{30.57}	&\textbf{56.95}	&9.50	&39.41	&\textbf{56.50}	&38.50	&\textbf{38.09}	&\textbf{79.50}	&\textbf{49.94}	&\textbf{49.57} \\
\bottomrule
\end{tabular}
\caption{Comparison across 12 benchmarks. The best is in bold. Abbreviations: M-Pro = MMLU-Pro, STEM = MMLU-STEM, W.G. = WinoGrande, H.S. = HellaSwag, BBH = Big-Bench.}
\label{tab:benchmark_results}
\end{table*}

\paragraph{Evaluation.} We adopt 12 benchmarks for comprehensive evaluation. Knowledge-intensive benchmarks include MMLU~\citep{hendrycks2021measuring}, CMMLU~\citep{li-etal-2024-cmmlu}, C-Eval~\citep{huang2023ceval}, MMLU-Pro~\citep{wang2024mmlupro}, and MMLU-STEM.  Mathematical capabilities are tested via GSM8K~\citep{cobbe2021training} and MATH~\citep{hendrycks2021measuringmath}. Reasoning abilities are measured using WinoGrande~\citep{sakaguchi2021winogrande}, HellaSwag~\citep{zellers-etal-2019-hellaswag}, ARC-C~\citep{clark2018think}, BIG-Bench~\citep{suzgun-etal-2023-challenging}, and DROP~\citep{dua-etal-2019-drop}.

\paragraph{Baselines.} We employ two baseline evaluation paradigms. The first assesses 40B-token general corpora, comprising the standard pre-training dataset and the web-sourced educational corpora FineWeb-Edu~\citep{penedo2024fineweb}, alongside KnowEdu, our curated high-quality knowledge-rich and educational data. 
The second paradigm assesses the QA blend following a 1:1 mixing ratio between KnowEdu and QA datasets.
General baselines include Nemotron-CC~\citep{su2024nemotron}, a blend of document and synthetic QA with a ratio of 9:1, and YulanQA, a QA subset extracted from the continual pre-training dataset of~\citet{chen2024towards}.
Mathematical baselines incorporate Nemotron-MIND~\citep{akter2025mind}, a dataset of synthetic math dialogues; MegaMathQA, a QA subset derived from MegaMath-Synthetic~\citep{zhou2025megamath}; and JiuZhang3.0~\citep{zhou2024jiuzhang}, a dataset of structured math problems with chain-of-thought (CoT). Dataset information is detailed in Table~\ref{tab:appendix_datasets} in Appendix~\ref{sec:appendix_datasets}.

\subsection{Main Results}

The main experimental results are shown in Table~\ref{tab:benchmark_results}, with mathematical results in Table~\ref{tab:mathresult}, from which we find that:

\paragraph{LinkQA achieves significant superiority over the general corpus baselines.} Compared to the pre-training baseline, LinkQA achieves an average improvement of \textbf{11.51\%} on MMLU and CMMLU, and 6.45\% across all benchmarks. LinkQA also demonstrates advantages over FineWeb-Edu and KnowEdu, with average improvements of 6.98\% and 4.85\%, respectively.

\paragraph{LinkQA boosts the performance across the vast majority of benchmarks, establishing a SOTA average.} LinkQA demonstrates the best on knowledge-intensive benchmarks. However, on the mathematical benchmarks GSM8K and Math, LinkQA lags slightly behind, probably due to the absence of CoT reasoning settings. For the average performance across all benchmarks, LinkQA outperforms the suboptimal JiuZhang3.0 by 1.56\%, demonstrating the advantage of our LinkSyn method.

\begin{figure}[t]
    \centering
    \includegraphics[width=0.9\columnwidth]{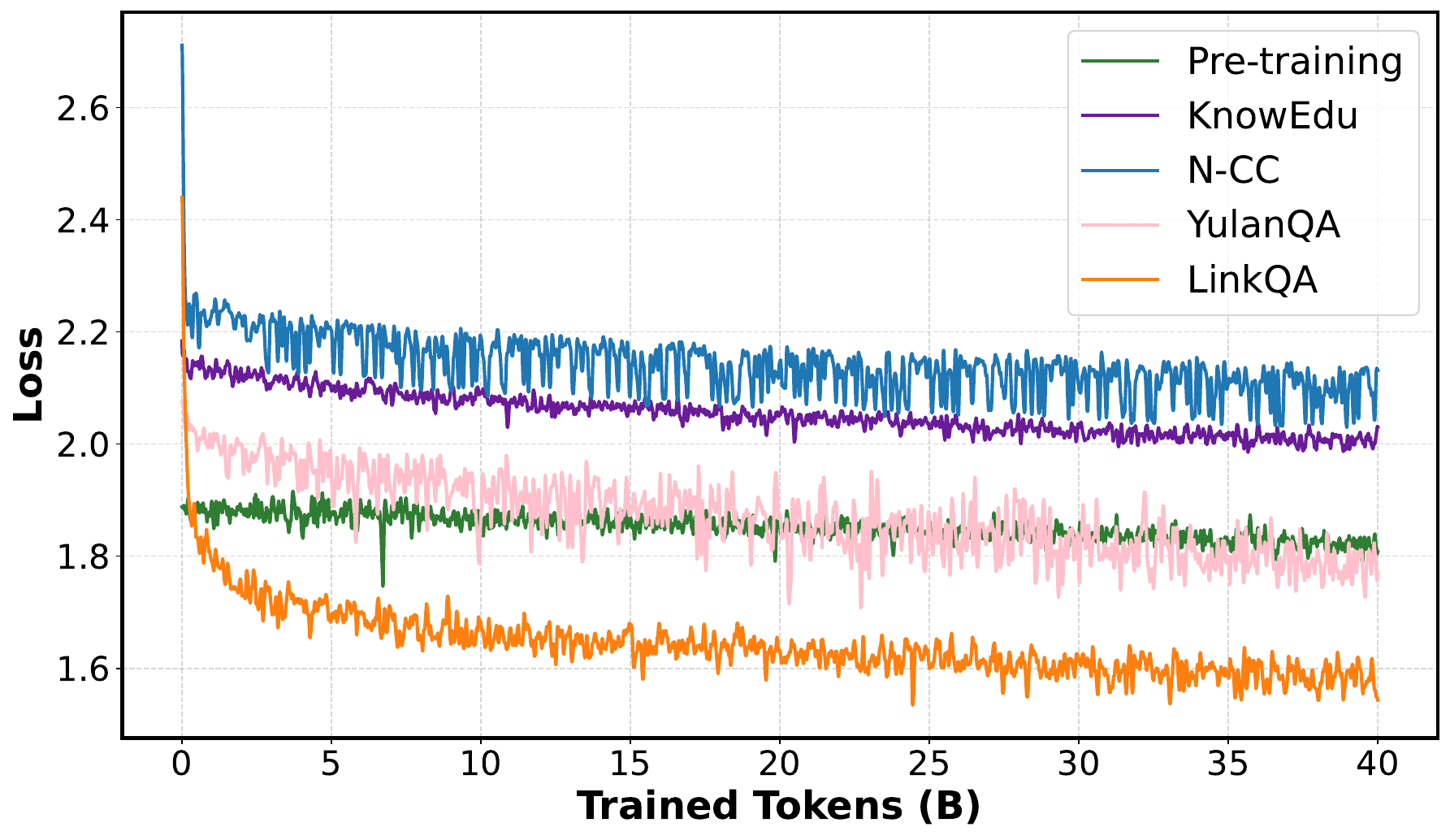}
    \caption{Training loss of LinkQA and baselines.}
    \label{fig:loss_comparison}
\end{figure}

\paragraph{LinkQA demonstrates sustained leading advantages during training.} Figure~\ref{fig:benchmark_comparison} reveals an expanding performance gap between LinkQA and the baselines as training progresses, particularly evident at 20B tokens. This indicates that LinkQA can continuously provide high-quality knowledge signals to models, promoting knowledge accumulation and integration while delivering long-term capability enhancement. As illustrated in Figure~\ref{fig:loss_comparison}, the loss changes during training show that the loss on LinkQA decreases at a rapid rate from the beginning and maintains lower loss values, which is consistent with the loss-performance correlation noted by~\citet{du2025understandingemergentabilitieslanguage}.
\begin{table}
    \centering
    \setlength{\tabcolsep}{1mm}
    \small
    \begin{tabular}{lcccccccc}
        \toprule
        \textbf{Dataset} & \textbf{CoT} & \textbf{G.} & \textbf{M.} & \textbf{Elem.} & \textbf{High.} & \textbf{Coll.} & \textbf{AVG.} \\
        \midrule
        Pre-training & - & 33.95 & 6.50 & 36.50 & 30.50 & 25.00 & 26.49 \\
        FineWeb-Edu & - & 31.49 & 2.50 & 38.00 & 31.00 & 30.00 & 26.60 \\
        KnowEdu & - & 32.56 & 8.00 & 37.50 & 27.50 &31.00 & 27.31 \\
        \midrule
        N-MIND & \CheckmarkBold & 47.42 & 13.50 & 43.50 & 29.50 & 37.00 & 34.18 \\
        MegaMathQA & \CheckmarkBold & 44.65 & 6.50 & 47.00 & 31.50 & 35.00 & 32.93 \\
        JiuZhang3.0 & \CheckmarkBold & \underline{56.27} & \textbf{23.00} & 42.50 & 30.50 & 31.00 & 36.65\\
        \midrule
        LinkQA & \XSolidBrush & 39.41 & 9.50 & 44.50 & 38.00 & \underline{44.00} & 35.08  \\ 
        LinkQA$_\text{Math}$ & \XSolidBrush & 42.96 & 13.50 & \textbf{57.00} & \textbf{46.50} & \textbf{46.00} & \underline{41.19} \\
        LinkQA$_\text{MathCoT}$ & \CheckmarkBold & \textbf{61.12} & \underline{21.50} & \underline{53.50} & \underline{44.50} & 38.00 & \textbf{43.72} \\
        \bottomrule
    \end{tabular}
    \caption{Comparison of mathematical performance. The best and second best are in bold and underlined, respectively. Abbreviations: G. = GSM8K, M. = MATH, Elem. = MMLU: elementary-mathematics, High. = MMLU: high-school-mathematics, Coll. = MMLU: college-mathematics.}
    \label{tab:mathresult}
\end{table}

\paragraph{Across mathematical benchmarks, LinkQA$_\text{MathCoT}$ improves by 7.07\% on average over the strongest baseline and achieves SOTA average performance.} As presented in Table~\ref{tab:mathresult}, LinkQA$_\text{MathCoT}$ outperforms JiuZhang3.0 by 4.85\% on GSM8K. LinkQA$_\text{Math}$ demonstrates exceptional performance on mathematical subsets of MMLU and significantly surpasses the second-best JiuZhang3.0 by 4.54\% on average. These highlight the effectiveness of LinkQA for comprehensive mathematical tasks.

\subsection{Scaling Analysis}
\begin{figure*}
    \centering
    \includegraphics[width=0.9\textwidth]{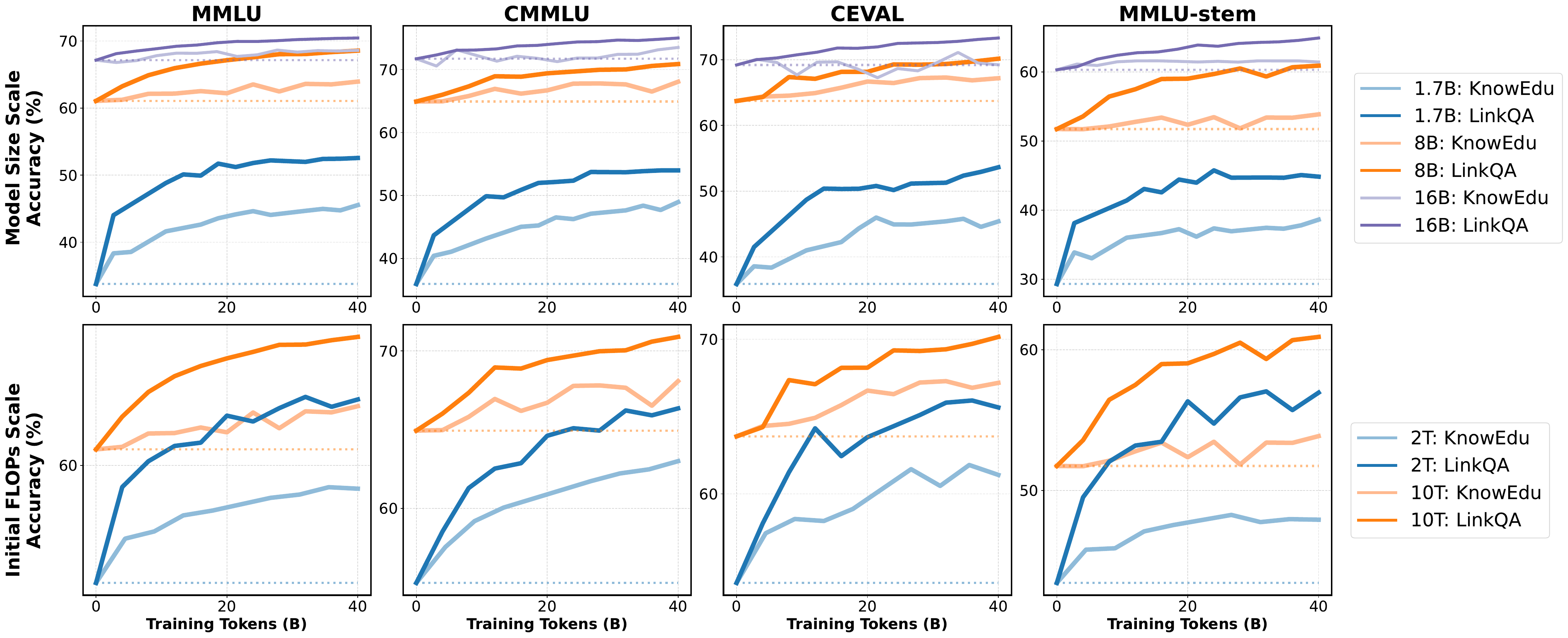}
    \caption{Scaling analysis of LinkQA across model and computational dimensions (detailed values in Appendix~\ref{sec:appendix_scaling}). For model size scalability, we use initial checkpoints of 4T, 10T, and 10T tokens for 1.7B, 8B, and 16B parameter models, respectively.}
    \label{fig:scaling_analysis}
\end{figure*}
We conduct scaling analysis experiments to validate the robustness of LinkQA, as shown in Figure~\ref{fig:scaling_analysis}.

\paragraph{LinkQA brings significant performance improvements across different model sizes.} For models with sizes of 1.7B, 8B, and 16B, as training progresses, the accuracy of LinkQA across different benchmarks consistently outperforms that of KnowEdu. This consistent enhancement across various model sizes demonstrates the excellent quality and generalization capability of LinkQA.

\paragraph{LinkQA consistently improves model performance regardless of initial FLOPs scale.} Different initial FLOPs reflect varying initial model capability at the start of continual pre-training. At both the 2T-token checkpoint and the 10T-token checkpoint, LinkQA demonstrates superior performance compared to KnowEdu. As training progresses, LinkQA exhibits an upward trend. This consistent performance across different pre-training stages with varying initial model capabilities further validates the high-quality characteristics of LinkQA.
\subsection{Ablation Studies}


\begin{table}[t]
\centering
\setlength{\tabcolsep}{1mm}
\small
\begin{tabular}{l|cccccccccc}
\toprule
\textbf{Dataset} & \textbf{MMLU}  & \textbf{C-Eval} & \textbf{ARC-C} & \textbf{DROP} & \textbf{AVG.} \\
\midrule
$\alpha=1$            & 59.40  & 61.43 & 71.50 & 44.48 & 59.20 \\
$\alpha=0$            & 58.96  & 61.40 & 73.50 & 46.04 & 59.98 \\
$\alpha=0.5$ & 59.94  & 62.89 & 74.00 & 47.01 & 60.96 \\
\midrule
$l=1$  & 58.91 & 61.79 & 72.00 & 44.30 & 59.25 \\
$l=2$    & 59.73 & 62.46 & 73.50& 46.89 & 60.65 \\
$l=3$  & 59.88 & 63.04 & 74.50 & 46.92 & 61.09 \\
\bottomrule
\end{tabular}
\caption{Comparison of different sampling policies ($\alpha \in \{0, 0.5, 1\}$) and different walk lengths ($l \in \{1, 2, 3\}$).}
\label{tab:sampling_strategy_ablation}
\end{table}
For sampling policy ablation, we test $\alpha=\{1,0.5,0\}$ (Eq.~\ref{eq:hybrid_equation}), while maintaining fixed ratios of each $l\in\{1,2,3\}$. For random walk length ablation, we fix $\alpha$  = 0.5 and synthesize datasets using exclusively $l\in\{1,2,3\}$ (Section~\ref{para:knowledge_value_optimization}). In both experiments, we combine the 4B-token LinkQA with 12B-token KnowEdu.

\paragraph{For sampling policies, the hybrid $\alpha=0.5$ achieves superior performance.} As shown in Table~\ref{tab:sampling_strategy_ablation}, $\alpha=0.5$ improves by 1.76\% compared to $\alpha=1$ and by 0.98\% compared to $\alpha=0$ on average, verifying the effectiveness of combining coverage and popularity sampling. For pure strategy comparison, in knowledge-intensive tasks such as MMLU and C-Eval, $\alpha=1$ shows a relative advantage over $\alpha=0$, indicating that KP coverage is more important for this type of task. Conversely, in complex reasoning tasks such as ARC-C and DROP, $\alpha=0$ is preferable, emphasizing the importance of knowledge popularity for such tasks.



\paragraph{For random walk length, increasing the length consistently improves the performance.} As shown in Table~\ref{tab:sampling_strategy_ablation}, the $l=3$ random walk achieves an average improvement of 1.84\% over $l=1$ and 0.44\% over $l=2$. This demonstrates that increasing random walk length in KP graph sampling effectively enhances the diversity and coverage of synthesized training data, leading to notable performance improvements, particularly for benchmarks focused on complex and compositional reasoning. However, it is important to note that longer walks also increase data synthesis costs, emphasizing the need for a practical trade-off between diversity and efficiency in large-scale data generation.

\subsection{Synthetic Data Analysis}
\begin{figure}[t]
    \centering
    \includegraphics[width=0.95\columnwidth]{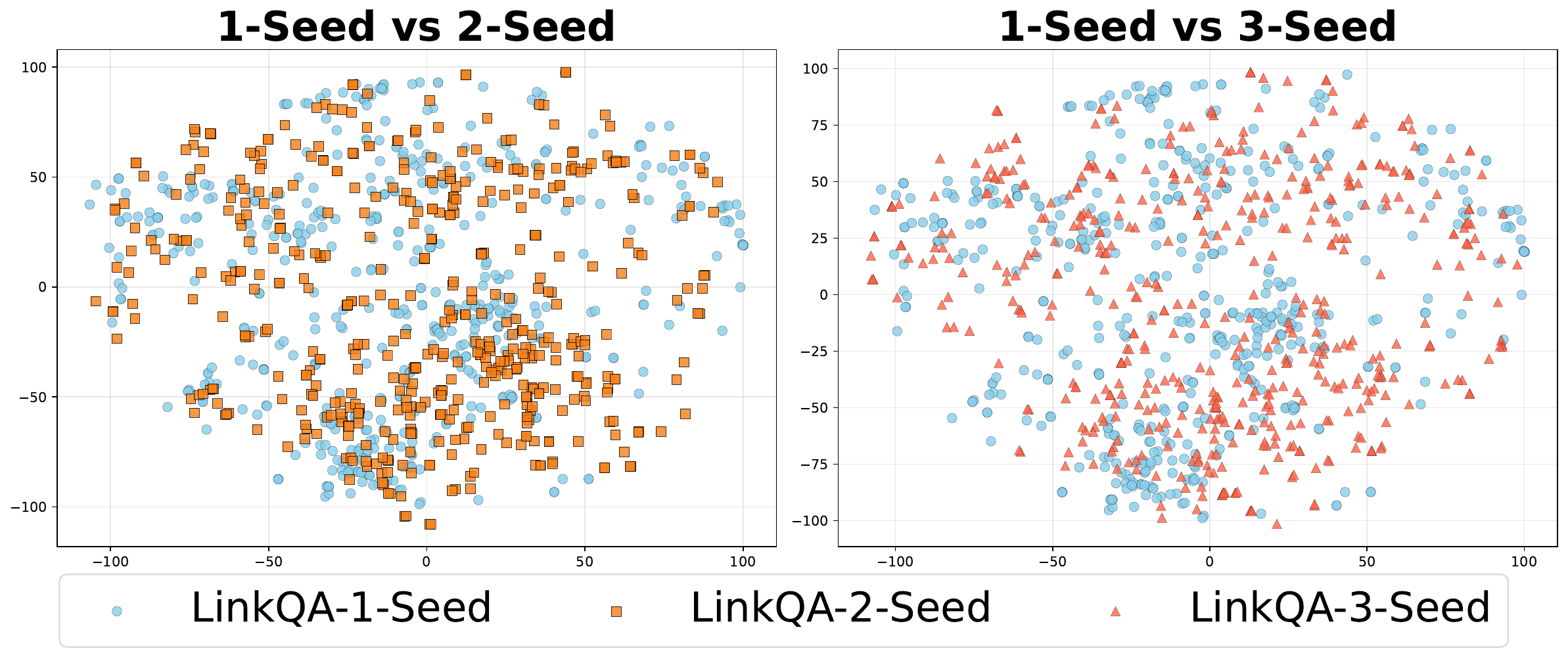}
    \caption{t-SNE visualization of semantic offsets between generated QA and seed data embeddings} 
    \label{fig:data_distribution_tsne}
\end{figure}

\begin{figure}[t]
    \centering
    \includegraphics[width=0.95\columnwidth]{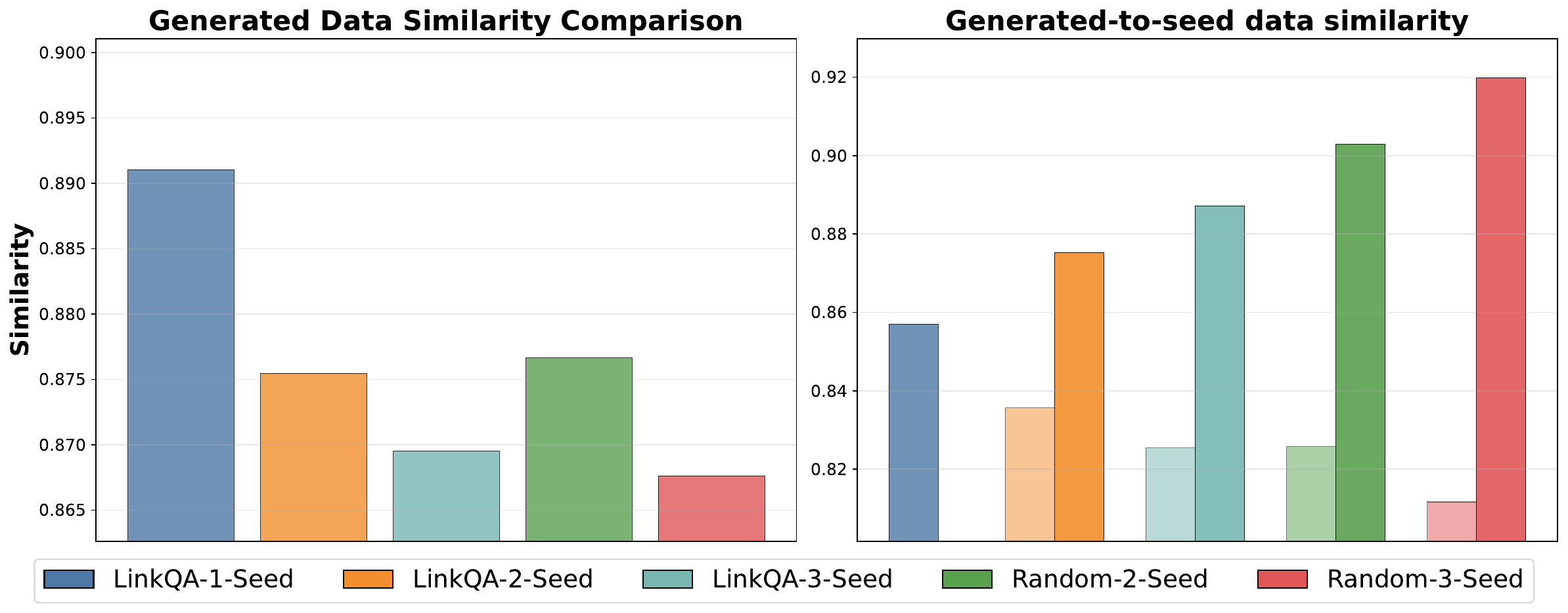}
    \caption{
    Left: Mean pairwise similarity among generated QA. 
    Right: Minimum and maximum similarity between generated and seed data.
    }
    \label{fig:similarity_of_generated_data}
\end{figure}

\begin{figure}[t]
    \centering
    \includegraphics[width=0.95\columnwidth]{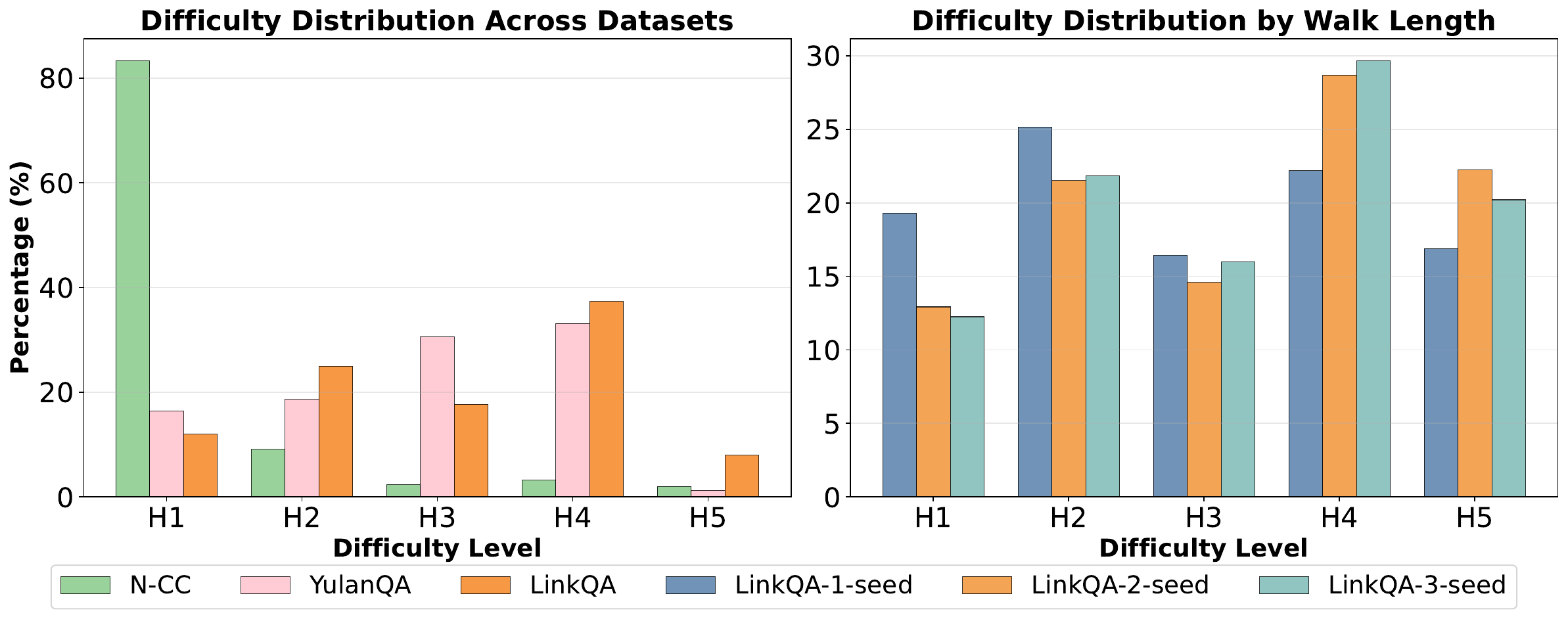}
    \caption{Left: Difficulty distribution of LinkQA vs. baseline. Right: Difficulty of 1/2/3-seed generated data with uniform seed difficulty.}
    \label{fig:Combined_Difficulty_Distribution_Comparison}
\end{figure}

We present a multi-dimensional analysis of synthesized data, including semantic diversity and distribution analysis.  For each setting ($l=1,2,3$), we sample 10,000 groups $(S_q, G_q)$, where $S_q = \{s_{q_1}, \ldots, s_{q_k}\}$ ($k=1,2,3$) are seed data, and $G_q = \{g_{q1}, \ldots, g_{qm}\}$ ($m=10$) are the corresponding generated data. To establish meaningful comparisons, we create control groups using randomly paired seeds (random-2/3-seed) without KP graph guidance. All results are visualized in Figures~\ref{fig:data_distribution_tsne},~\ref{fig:similarity_of_generated_data}, and~\ref{fig:Combined_Difficulty_Distribution_Comparison}.

\paragraph{Multi-seed synthesis achieves broader and more uniform semantic diffusion than single-seed generation.} We use Sentence-T5~\citep{ni2021sentencet5scalablesentenceencoders} to embed the sampled data. First, we compute the offset vectors of generated data relative to their seed data. Figure~\ref{fig:data_distribution_tsne} shows the t-SNE visualization of 500 such offsets per group, demonstrating that 2/3-seed distributions cover a larger and more uniform semantic space. Next, we calculate the mean cosine similarity among the generated data within each group: $\mathrm{mean}_q\,\mathrm{mean}_{i \neq j}\,\cos(g_{qi}, g_{qj})$. As shown in the left panel of Figure~\ref{fig:similarity_of_generated_data}, using more seeds results in lower similarity and thus greater diversity among the generated data.

\paragraph{KP graph-based sampling enables effective semantic fusion across seeds.} To evaluate semantic integration, we compute $\mathrm{mean}_q\,\mathrm{mean}_i\,\operatorname{agg}_{s \in S_q} \operatorname{sim}(g_{qi}, s)$, where $\operatorname{agg}$ is either $\max$ or $\min$, representing the maximum or minimum similarity between each generated data and its corresponding seeds. As shown in the right panel of Figure~\ref{fig:similarity_of_generated_data}, graph-based multi-seed generation yields a much smaller gap (0.04/0.06) than random sampling (0.07/0.11), with comparable overall diversity. This indicates that graph-based sampling effectively fuses semantics across seeds, while random sampling produces examples closely tied to a single seed.

\paragraph{LinkQA contains a higher proportion of high-difficulty data than baselines, and multi-seed synthesis further increases this proportion.} The left panel of Figure~\ref{fig:Combined_Difficulty_Distribution_Comparison} illustrates the difficulty distributions of LinkQA and baselines, based on 100{,}000 randomly sampled and annotated instances from each dataset. LinkQA includes approximately $10\times$ more high-difficulty items (H4, H5) than Nemotron-CC. Notably, it even surpasses YulanQA, which, although not synthetic, is filtered for challenging QA yet still contains fewer high‑difficulty items. The right panel demonstrates that multi-seed synthesis further elevates the proportion of challenging questions, with 2/3-seed methods yielding 10\% more high-difficulty items than the 1-seed method, indicating that integrating multiple knowledge sources enhances question complexity. Detailed distribution analysis of LinkQA is provided in Appendix~\ref{sec:LinkQA_distribution}.

\subsection{Case Study}
We further conduct a case study to evaluate the quality and accuracy of LinkQA, as detailed in Appendix~\ref{sec:Case_Study} and Appendix~\ref{sec:appendix_qualityreview}. The QA pairs generated with varying difficulty levels, disciplines, question types, and seed data counts demonstrate both the multi-dimensional diversity and the high quality of our synthesis pipeline.

\section{Related Work}
Existing work on pre‑training data synthesis for LLMs has produced a diverse range of corpora and methods. General corpora such as FineWeb‑Edu~\citep{penedo2024fineweb} provide broad coverage but lack explicit QA supervision. In contrast, QA data shows superior performance~\citep{wang2025octothinkermidtrainingincentivizesreinforcement}. Several studies aim to improve QA quality: \citet{maini2024rephrasingwebrecipecompute} rephrases pre‑training corpora into QA form; \citet{cheng2024instruction} designs instruction‑driven synthesis; and \citet{jiang2025mixcpt} evaluates QA integration in continual pre‑training. Large‑scale pipelines include Nemotron~\citep{su2024nemotron}, which generates 499.5B‑token of document–QA pairs; MIND~\citep{akter2025mind}, which creates 138B‑token of role‑specific math dialogues; MegaMath~\citep{zhou2025megamath}, a 7B‑token dataset refined from mathematics‑related webpages; and JiuZhang3.0~\citep{zhou2024jiuzhang}, a 4.6B‑token distilled corpus for mathematical QA. These methods show promise for synthetic QA but face challenges in scaling reasoning quality, ensuring concept diversity, and supporting multi‑domain generalization.

To address these gaps, recent methods have adopted graph-based sampling to introduce knowledge structure. Entity-graph approaches~\citep{qin2025scalinglawssyntheticdata,Jiang2025SynthesizeonGraphKSA} link texts via co-occurring entities, but entities often reflect surface mentions rather than the underlying concepts being examined. In contrast, we build graphs over knowledge points to capture tighter logical relations and enable control over difficulty, discipline, and KP distribution. This forms the basis of LinkSyn, through which we generate LinkQA, achieving SOTA results on 12 benchmarks.

\section{Conclusion}

In this paper, we introduce LinkSyn, a novel KP graph-based synthesis framework. By extracting KPs from QA seed data and constructing KP graphs, LinkSyn performs diffusion-based QA synthesis via DeepSeek-R1, based on multiple seeds that are strongly linked by KPs and sampled from graph walks. The synthesized LinkQA dataset significantly advances multi-disciplinary capabilities, as demonstrated by an 11.51\% average improvement on MMLU and CMMLU when continually pre-training Llama-3 8B. These SOTA results, coupled with consistent gains across model size and initial FLOPs scales, underscore LinkSyn’s efficacy in generating diverse, valuable synthetic QA data.

\bibliography{aaai2026}

\begin{thebibliography}{45}
\providecommand{\natexlab}[1]{#1}

\bibitem[{Akter et~al.(2025)Akter, Prabhumoye, Kamalu, Satheesh, Nyberg, Patwary, Shoeybi, and Catanzaro}]{akter2025mind}
Akter, S.~N.; Prabhumoye, S.; Kamalu, J.; Satheesh, S.; Nyberg, E.; Patwary, M.; Shoeybi, M.; and Catanzaro, B. 2025.
\newblock {MIND}: Math Informed syNthetic Dialogues for Pretraining {LLM}s.
\newblock In \emph{The Thirteenth International Conference on Learning Representations}.

\bibitem[{Chen et~al.(2025)Chen, Chen, Wang, Zhou, Zhu, Jiang, Min, Zhao, Dou, Mao, Lin, Song, Xu, Chen, Yan, Wei, Hu, Huang, and Wen}]{chen2024towards}
Chen, J.; Chen, Z.; Wang, J.; Zhou, K.; Zhu, Y.; Jiang, J.; Min, Y.; Zhao, X.; Dou, Z.; Mao, J.; Lin, Y.; Song, R.; Xu, J.; Chen, X.; Yan, R.; Wei, Z.; Hu, D.; Huang, W.; and Wen, J.-R. 2025.
\newblock Towards Effective and Efficient Continual Pre-training of Large Language Models.
\newblock In \emph{Proceedings of the 63rd Annual Meeting of the Association for Computational Linguistics (Volume 1: Long Papers)}, 5779--5795.

\bibitem[{Cheng et~al.(2024)Cheng, Gu, Huang, Bi, Huang, and Wei}]{cheng2024instruction}
Cheng, D.; Gu, Y.; Huang, S.; Bi, J.; Huang, M.; and Wei, F. 2024.
\newblock Instruction Pre-Training: Language Models are Supervised Multitask Learners.
\newblock In \emph{Proceedings of the 2024 Conference on Empirical Methods in Natural Language Processing}, 2529--2550.

\bibitem[{Clark et~al.(2018)Clark, Cowhey, Etzioni, Khot, Sabharwal, Schoenick, and Tafjord}]{clark2018think}
Clark, P.; Cowhey, I.; Etzioni, O.; Khot, T.; Sabharwal, A.; Schoenick, C.; and Tafjord, O. 2018.
\newblock Think you have Solved Question Answering? Try ARC, the {AI2} Reasoning Challenge.
\newblock \emph{CoRR}, abs/1803.05457.

\bibitem[{Cobbe et~al.(2021)Cobbe, Kosaraju, Bavarian, Chen, Jun, Kaiser, Plappert, Tworek, Hilton, Nakano, Hesse, and Schulman}]{cobbe2021training}
Cobbe, K.; Kosaraju, V.; Bavarian, M.; Chen, M.; Jun, H.; Kaiser, L.; Plappert, M.; Tworek, J.; Hilton, J.; Nakano, R.; Hesse, C.; and Schulman, J. 2021.
\newblock Training Verifiers to Solve Math Word Problems.
\newblock arXiv:2110.14168.

\bibitem[{DeepSeek-AI et~al.(2025)DeepSeek-AI, Guo, Yang, Zhang, Song, Zhang, Xu, Zhu, Ma, Wang, Bi, Zhang, Yu, Wu, Wu, Gou, Shao, Li, Gao, Liu, Xue, Wang, Wu, Feng, Lu, Zhao, Deng, Zhang, Ruan, Dai, Chen, Ji, Li, Lin, Dai, Luo, Hao, Chen, Li, Zhang, Bao, Xu, Wang, Ding, Xin, Gao, Qu, Li, Guo, Li, Wang, Chen, Yuan, Qiu, Li, Cai, Ni, Liang, Chen, Dong, Hu, Gao, Guan, Huang, Yu, Wang, Zhang, Zhao, Wang, Zhang, Xu, Xia, Zhang, Zhang, Tang, Li, Wang, Li, Tian, Huang, Zhang, Wang, Chen, Du, Ge, Zhang, Pan, Wang, Chen, Jin, Chen, Lu, Zhou, Chen, Ye, Wang, Yu, Zhou, Pan, and Li}]{deepseekai2025deepseekr1incentivizingreasoningcapability}
DeepSeek-AI; Guo, D.; Yang, D.; Zhang, H.; Song, J.; Zhang, R.; Xu, R.; Zhu, Q.; Ma, S.; Wang, P.; Bi, X.; Zhang, X.; Yu, X.; Wu, Y.; Wu, Z.~F.; Gou, Z.; Shao, Z.; Li, Z.; Gao, Z.; Liu, A.; Xue, B.; Wang, B.; Wu, B.; Feng, B.; Lu, C.; Zhao, C.; Deng, C.; Zhang, C.; Ruan, C.; Dai, D.; Chen, D.; Ji, D.; Li, E.; Lin, F.; Dai, F.; Luo, F.; Hao, G.; Chen, G.; Li, G.; Zhang, H.; Bao, H.; Xu, H.; Wang, H.; Ding, H.; Xin, H.; Gao, H.; Qu, H.; Li, H.; Guo, J.; Li, J.; Wang, J.; Chen, J.; Yuan, J.; Qiu, J.; Li, J.; Cai, J.~L.; Ni, J.; Liang, J.; Chen, J.; Dong, K.; Hu, K.; Gao, K.; Guan, K.; Huang, K.; Yu, K.; Wang, L.; Zhang, L.; Zhao, L.; Wang, L.; Zhang, L.; Xu, L.; Xia, L.; Zhang, M.; Zhang, M.; Tang, M.; Li, M.; Wang, M.; Li, M.; Tian, N.; Huang, P. et~al. 2025.
\newblock DeepSeek-R1: Incentivizing Reasoning Capability in LLMs via Reinforcement Learning.
\newblock \emph{CoRR}, abs/2501.12948.

\bibitem[{DeepSeek-AI et~al.(2024)DeepSeek-AI, Liu, Feng, Xue, Wang, Wu, Lu, Zhao, Deng, Zhang, Ruan, Dai, Guo, Yang, Chen, Ji, Li, Lin, Dai, Luo, Hao, Chen, Li, Zhang, Bao, Xu, Wang, Zhang, Ding, Xin, Gao, Li, Qu, Cai, Liang, Guo, Ni, Li, Wang, Chen, Chen, Yuan, Qiu, Li, Song, Dong, Hu, Gao, Guan, Huang, Yu, Wang, Zhang, Xu, Xia, Zhao, Wang, Zhang, Li, Wang, Zhang, Zhang, Tang, Li, Tian, Huang, Wang, Zhang, Wang, Zhu, Chen, Du, Chen, Jin, Ge, Zhang, Pan, Wang, Xu, Zhang, Chen, Li, Lu, Zhou, Chen, Wu, Ye, Ye, Ma, Wang, Zhou, Yu, Zhou, Pan, Wang, Yun, Pei, Sun, Xiao, and Zeng}]{deepseekai2025deepseekv3technicalreport}
DeepSeek-AI; Liu, A.; Feng, B.; Xue, B.; Wang, B.; Wu, B.; Lu, C.; Zhao, C.; Deng, C.; Zhang, C.; Ruan, C.; Dai, D.; Guo, D.; Yang, D.; Chen, D.; Ji, D.; Li, E.; Lin, F.; Dai, F.; Luo, F.; Hao, G.; Chen, G.; Li, G.; Zhang, H.; Bao, H.; Xu, H.; Wang, H.; Zhang, H.; Ding, H.; Xin, H.; Gao, H.; Li, H.; Qu, H.; Cai, J.~L.; Liang, J.; Guo, J.; Ni, J.; Li, J.; Wang, J.; Chen, J.; Chen, J.; Yuan, J.; Qiu, J.; Li, J.; Song, J.; Dong, K.; Hu, K.; Gao, K.; Guan, K.; Huang, K.; Yu, K.; Wang, L.; Zhang, L.; Xu, L.; Xia, L.; Zhao, L.; Wang, L.; Zhang, L.; Li, M.; Wang, M.; Zhang, M.; Zhang, M.; Tang, M.; Li, M.; Tian, N.; Huang, P.; Wang, P.; Zhang, P.; Wang, Q.; Zhu, Q.; Chen, Q.; Du, Q.; Chen, R.~J.; Jin, R.~L.; Ge, R.; Zhang, R.; Pan, R.; Wang, R.; Xu, R.; Zhang, R. et~al. 2024.
\newblock DeepSeek-V3 Technical Report.
\newblock \emph{CoRR}, abs/2412.19437.

\bibitem[{Du et~al.(2024)Du, Zeng, Dong, and Tang}]{du2025understandingemergentabilitieslanguage}
Du, Z.; Zeng, A.; Dong, Y.; and Tang, J. 2024.
\newblock Understanding Emergent Abilities of Language Models from the Loss Perspective.
\newblock In \emph{The Thirty-eighth Annual Conference on Neural Information Processing Systems}.

\bibitem[{Dua et~al.(2019)Dua, Wang, Dasigi, Stanovsky, Singh, and Gardner}]{dua-etal-2019-drop}
Dua, D.; Wang, Y.; Dasigi, P.; Stanovsky, G.; Singh, S.; and Gardner, M. 2019.
\newblock {DROP}: A Reading Comprehension Benchmark Requiring Discrete Reasoning Over Paragraphs.
\newblock In \emph{Proceedings of the 2019 Conference of the North {A}merican Chapter of the Association for Computational Linguistics: Human Language Technologies, Volume 1 (Long and Short Papers)}, 2368--2378.

\bibitem[{Duan et~al.(2025)Duan, Zhang, Wang, Que, Liu, Rong, and Cai}]{duan2025enhancing}
Duan, F.; Zhang, X.; Wang, S.; Que, H.; Liu, Y.; Rong, W.; and Cai, X. 2025.
\newblock Enhancing llms via high-knowledge data selection.
\newblock In \emph{Proceedings of the AAAI Conference on Artificial Intelligence}, 23832--23840.

\bibitem[{Dubey et~al.(2024)Dubey, Jauhri, Pandey, Kadian, Al-Dahle, Letman, Mathur, Schelten, Yang, Fan, Goyal, Hartshorn, Yang, Mitra, Sravankumar, Korenev, Hinsvark, Rao, Zhang, Rodriguez, Gregerson, Spataru, Rozière, Biron, Tang, Chern, Caucheteux, Nayak, Bi, Marra, McConnell, Keller, Touret, Wu, Wong, Ferrer, Nikolaidis, Allonsius, Song, Pintz, Livshits, Esiobu, Choudhary, Mahajan, Garcia-Olano, Perino, Hupkes, Lakomkin, AlBadawy, Lobanova, Dinan, Smith, Radenovic, Zhang, Synnaeve, Lee, Anderson, Nail, Mialon, Pang, Cucurell, Nguyen, Korevaar, Xu, Touvron, Zarov, Ibarra, Kloumann, Misra, Evtimov, Copet, Lee, Geffert, Vranes, Park, Mahadeokar, Shah, van~der Linde, Billock, Hong, Lee, Fu, Chi, Huang, Liu, Wang, Yu, Bitton, Spisak, Park, Rocca, Johnstun, Saxe, Jia, Alwala, Upasani, Plawiak, Li, Heafield, Stone, and et~al.}]{grattafiori2024llama3herdmodels}
Dubey, A.; Jauhri, A.; Pandey, A.; Kadian, A.; Al-Dahle, A.; Letman, A.; Mathur, A.; Schelten, A.; Yang, A.; Fan, A.; Goyal, A.; Hartshorn, A.; Yang, A.; Mitra, A.; Sravankumar, A.; Korenev, A.; Hinsvark, A.; Rao, A.; Zhang, A.; Rodriguez, A.; Gregerson, A.; Spataru, A.; Rozière, B.; Biron, B.; Tang, B.; Chern, B.; Caucheteux, C.; Nayak, C.; Bi, C.; Marra, C.; McConnell, C.; Keller, C.; Touret, C.; Wu, C.; Wong, C.; Ferrer, C.~C.; Nikolaidis, C.; Allonsius, D.; Song, D.; Pintz, D.; Livshits, D.; Esiobu, D.; Choudhary, D.; Mahajan, D.; Garcia-Olano, D.; Perino, D.; Hupkes, D.; Lakomkin, E.; AlBadawy, E.; Lobanova, E.; Dinan, E.; Smith, E.~M.; Radenovic, F.; Zhang, F.; Synnaeve, G.; Lee, G.; Anderson, G.~L.; Nail, G.; Mialon, G.; Pang, G.; Cucurell, G.; Nguyen, H.; Korevaar, H.; Xu, H.; Touvron, H.; Zarov, I.; Ibarra, I.~A.; Kloumann, I.~M.; Misra, I.; Evtimov, I.; Copet, J.; Lee, J.; Geffert, J.; Vranes, J.; Park, J.; Mahadeokar, J.; Shah, J.; van~der Linde, J.; Billock, J.; Hong, J. et~al. 2024.
\newblock The Llama 3 Herd of Models.
\newblock \emph{CoRR}, abs/2407.21783.

\bibitem[{Gunasekar et~al.(2023)Gunasekar, Zhang, Aneja, Mendes, Giorno, Gopi, Javaheripi, Kauffmann, de~Rosa, Saarikivi, Salim, Shah, Behl, Wang, Bubeck, Eldan, Kalai, Lee, and Li}]{gunasekar2023textbooksneed}
Gunasekar, S.; Zhang, Y.; Aneja, J.; Mendes, C. C.~T.; Giorno, A.~D.; Gopi, S.; Javaheripi, M.; Kauffmann, P.; de~Rosa, G.; Saarikivi, O.; Salim, A.; Shah, S.; Behl, H.~S.; Wang, X.; Bubeck, S.; Eldan, R.; Kalai, A.~T.; Lee, Y.~T.; and Li, Y. 2023.
\newblock Textbooks Are All You Need.
\newblock arXiv:2306.11644.

\bibitem[{Hao et~al.(2022)Hao, Gong, Yu, and Loia}]{HAO202296}
Hao, F.; Gong, Y.; Yu, W.; and Loia, V. 2022.
\newblock Knowledge points navigation based on three-way concept lattice for autonomous learning.
\newblock \emph{Pattern Recognition Letters}, 163: 96--103.

\bibitem[{Hendrycks et~al.(2021{\natexlab{a}})Hendrycks, Burns, Basart, Zou, Mazeika, Song, and Steinhardt}]{hendrycks2021measuringmassivemultitasklanguage}
Hendrycks, D.; Burns, C.; Basart, S.; Zou, A.; Mazeika, M.; Song, D.; and Steinhardt, J. 2021{\natexlab{a}}.
\newblock Measuring Massive Multitask Language Understanding.
\newblock In \emph{International Conference on Learning Representations}.

\bibitem[{Hendrycks et~al.(2021{\natexlab{b}})Hendrycks, Burns, Basart, Zou, Mazeika, Song, and Steinhardt}]{hendrycks2021measuring}
Hendrycks, D.; Burns, C.; Basart, S.; Zou, A.; Mazeika, M.; Song, D.; and Steinhardt, J. 2021{\natexlab{b}}.
\newblock Measuring Massive Multitask Language Understanding.
\newblock In \emph{International Conference on Learning Representations}.

\bibitem[{Hendrycks et~al.(2021{\natexlab{c}})Hendrycks, Burns, Kadavath, Arora, Basart, Tang, Song, and Steinhardt}]{hendrycks2021measuringmath}
Hendrycks, D.; Burns, C.; Kadavath, S.; Arora, A.; Basart, S.; Tang, E.; Song, D.; and Steinhardt, J. 2021{\natexlab{c}}.
\newblock Measuring Mathematical Problem Solving With the {MATH} Dataset.
\newblock In \emph{Thirty-fifth Conference on Neural Information Processing Systems Datasets and Benchmarks Track (Round 2)}.

\bibitem[{Huang et~al.(2024)Huang, Wen, Zhao, Hu, Liu, Jia, Mao, Wang, Zhang, Chen, Chen, and Zhang}]{huang2024subjectdrivescalinggenerativedata}
Huang, B.; Wen, Y.; Zhao, Y.; Hu, Y.; Liu, Y.; Jia, F.; Mao, W.; Wang, T.; Zhang, C.; Chen, C.~W.; Chen, Z.; and Zhang, X. 2024.
\newblock SubjectDrive: Scaling Generative Data in Autonomous Driving via Subject Control.
\newblock arXiv:2403.19438.

\bibitem[{Huang et~al.(2025)Huang, Yu, Ma, Zhong, Feng, Wang, Chen, Peng, Feng, Qin, and Liu}]{Huang_2025}
Huang, L.; Yu, W.; Ma, W.; Zhong, W.; Feng, Z.; Wang, H.; Chen, Q.; Peng, W.; Feng, X.; Qin, B.; and Liu, T. 2025.
\newblock A Survey on Hallucination in Large Language Models: Principles, Taxonomy, Challenges, and Open Questions.
\newblock \emph{ACM Transactions on Information Systems}, 1–55.

\bibitem[{Huang et~al.(2023)Huang, Bai, Zhu, Zhang, Zhang, Su, Liu, Lv, Zhang, jiayi lei, Fu, Sun, and He}]{huang2023ceval}
Huang, Y.; Bai, Y.; Zhu, Z.; Zhang, J.; Zhang, J.; Su, T.; Liu, J.; Lv, C.; Zhang, Y.; jiayi lei; Fu, Y.; Sun, M.; and He, J. 2023.
\newblock C-Eval: A Multi-Level Multi-Discipline Chinese Evaluation Suite for Foundation Models.
\newblock In \emph{Thirty-seventh Conference on Neural Information Processing Systems Datasets and Benchmarks Track}.

\bibitem[{Jiang et~al.(2025{\natexlab{a}})Jiang, Li, Zhao, Song, Zhang, and Wen}]{jiang2025mixcpt}
Jiang, J.; Li, J.; Zhao, X.; Song, Y.; Zhang, T.; and Wen, J.-R. 2025{\natexlab{a}}.
\newblock Mix-{CPT}: A Domain Adaptation Framework via Decoupling Knowledge Learning and Format Alignment.
\newblock In \emph{The Thirteenth International Conference on Learning Representations}.

\bibitem[{Jiang et~al.(2025{\natexlab{b}})Jiang, Ma, Xu, Yang, Zhang, and Guo}]{Jiang2025SynthesizeonGraphKSA}
Jiang, X.; Ma, S.; Xu, C.; Yang, C.; Zhang, L.; and Guo, J. 2025{\natexlab{b}}.
\newblock Synthesize-on-Graph: Knowledgeable Synthetic Data Generation for Continue Pre-training of Large Language Models.
\newblock arXiv:2505.00979.

\bibitem[{Kandpal et~al.(2023)Kandpal, Deng, Roberts, Wallace, and Raffel}]{kandpal2023largelanguagemodelsstruggle}
Kandpal, N.; Deng, H.; Roberts, A.; Wallace, E.; and Raffel, C. 2023.
\newblock Large Language Models Struggle to Learn Long-Tail Knowledge.
\newblock arXiv:2211.08411.

\bibitem[{Li et~al.(2024)Li, Zhang, Koto, Yang, Zhao, Gong, Duan, and Baldwin}]{li-etal-2024-cmmlu}
Li, H.; Zhang, Y.; Koto, F.; Yang, Y.; Zhao, H.; Gong, Y.; Duan, N.; and Baldwin, T. 2024.
\newblock {CMMLU}: Measuring massive multitask language understanding in {C}hinese.
\newblock In \emph{Findings of the Association for Computational Linguistics: ACL 2024}, 11260--11285.

\bibitem[{Li et~al.(2023)Li, Bubeck, Eldan, Giorno, Gunasekar, and Lee}]{li2023textbooksneediiphi15}
Li, Y.; Bubeck, S.; Eldan, R.; Giorno, A.~D.; Gunasekar, S.; and Lee, Y.~T. 2023.
\newblock Textbooks Are All You Need II: phi-1.5 technical report.
\newblock arXiv:2309.05463.

\bibitem[{Maini et~al.(2024)Maini, Seto, Bai, Grangier, Zhang, and Jaitly}]{maini2024rephrasingwebrecipecompute}
Maini, P.; Seto, S.; Bai, R.; Grangier, D.; Zhang, Y.; and Jaitly, N. 2024.
\newblock Rephrasing the Web: A Recipe for Compute and Data-Efficient Language Modeling.
\newblock In \emph{Proceedings of the 62nd Annual Meeting of the Association for Computational Linguistics (Volume 1: Long Papers)}, 14044--14072.

\bibitem[{Muennighoff et~al.(2025)Muennighoff, Rush, Barak, Scao, Piktus, Tazi, Pyysalo, Wolf, and Raffel}]{muennighoff2025scalingdataconstrainedlanguagemodels}
Muennighoff, N.; Rush, A.~M.; Barak, B.; Scao, T.~L.; Piktus, A.; Tazi, N.; Pyysalo, S.; Wolf, T.; and Raffel, C. 2025.
\newblock Scaling Data-Constrained Language Models.
\newblock arXiv:2305.16264.

\bibitem[{Nadăș, Dioșan, and Tomescu(2025)}]{nadas2025syntheticdatagenerationusing}
Nadăș, M.; Dioșan, L.; and Tomescu, A. 2025.
\newblock Synthetic Data Generation Using Large Language Models: Advances in Text and Code.
\newblock \emph{IEEE Access}, 1–1.

\bibitem[{Narayanan et~al.(2021)Narayanan, Shoeybi, Casper, LeGresley, Patwary, Korthikanti, Vainbrand, Kashinkunti, Bernauer, Catanzaro, Phanishayee, and Zaharia}]{shoeybi2020megatronlmtrainingmultibillionparameter}
Narayanan, D.; Shoeybi, M.; Casper, J.; LeGresley, P.; Patwary, M.; Korthikanti, V.; Vainbrand, D.; Kashinkunti, P.; Bernauer, J.; Catanzaro, B.; Phanishayee, A.; and Zaharia, M. 2021.
\newblock Efficient large-scale language model training on GPU clusters using megatron-LM.
\newblock In \emph{Proceedings of the International Conference for High Performance Computing, Networking, Storage and Analysis}, SC '21. New York, NY, USA: Association for Computing Machinery.
\newblock ISBN 9781450384421.

\bibitem[{Ni et~al.(2021)Ni, Ábrego, Constant, Ma, Hall, Cer, and Yang}]{ni2021sentencet5scalablesentenceencoders}
Ni, J.; Ábrego, G.~H.; Constant, N.; Ma, J.; Hall, K.~B.; Cer, D.; and Yang, Y. 2021.
\newblock Sentence-T5: Scalable Sentence Encoders from Pre-trained Text-to-Text Models.
\newblock \emph{arXiv preprint arXiv:2108.08877}.

\bibitem[{Penedo et~al.(2024)Penedo, Kydl{\'\i}{\v{c}}ek, allal, Lozhkov, Mitchell, Raffel, Werra, and Wolf}]{penedo2024fineweb}
Penedo, G.; Kydl{\'\i}{\v{c}}ek, H.; allal, L.~B.; Lozhkov, A.; Mitchell, M.; Raffel, C.; Werra, L.~V.; and Wolf, T. 2024.
\newblock The FineWeb Datasets: Decanting the Web for the Finest Text Data at Scale.
\newblock In \emph{The Thirty-eight Conference on Neural Information Processing Systems Datasets and Benchmarks Track}.

\bibitem[{Qin et~al.(2025)Qin, Dong, Zhang, Dong, Huang, Yang, Khademi, Zhang, Awadalla, Fung, Chen, Cheng, and Wei}]{qin2025scalinglawssyntheticdata}
Qin, Z.; Dong, Q.; Zhang, X.; Dong, L.; Huang, X.; Yang, Z.; Khademi, M.; Zhang, D.; Awadalla, H.~H.; Fung, Y.~R.; Chen, W.; Cheng, M.; and Wei, F. 2025.
\newblock Scaling Laws of Synthetic Data for Language Models.
\newblock arXiv:2503.19551.

\bibitem[{Qwen et~al.(2025)Qwen, :, Yang, Yang, Zhang, Hui, Zheng, Yu, Li, Liu, Huang, Wei, Lin, Yang, Tu, Zhang, Yang, Yang, Zhou, Lin, Dang, Lu, Bao, Yang, Yu, Li, Xue, Zhang, Zhu, Men, Lin, Li, Tang, Xia, Ren, Ren, Fan, Su, Zhang, Wan, Liu, Cui, Zhang, and Qiu}]{qwen2025qwen25technicalreport}
Qwen; :; Yang, A.; Yang, B.; Zhang, B.; Hui, B.; Zheng, B.; Yu, B.; Li, C.; Liu, D.; Huang, F.; Wei, H.; Lin, H.; Yang, J.; Tu, J.; Zhang, J.; Yang, J.; Yang, J.; Zhou, J.; Lin, J.; Dang, K.; Lu, K.; Bao, K.; Yang, K.; Yu, L.; Li, M.; Xue, M.; Zhang, P.; Zhu, Q.; Men, R.; Lin, R.; Li, T.; Tang, T.; Xia, T.; Ren, X.; Ren, X.; Fan, Y.; Su, Y.; Zhang, Y.; Wan, Y.; Liu, Y.; Cui, Z.; Zhang, Z.; and Qiu, Z. 2025.
\newblock Qwen2.5 Technical Report.
\newblock arXiv:2412.15115.

\bibitem[{Sakaguchi et~al.(2021)Sakaguchi, Bras, Bhagavatula, and Choi}]{sakaguchi2021winogrande}
Sakaguchi, K.; Bras, R.~L.; Bhagavatula, C.; and Choi, Y. 2021.
\newblock Winogrande: An adversarial winograd schema challenge at scale.
\newblock \emph{Communications of the ACM}, 64(9): 99--106.

\bibitem[{Shao et~al.(2024)Shao, Wang, Zhu, Xu, Song, Bi, Zhang, Zhang, Li, Wu, and Guo}]{shao2024deepseekmathpushinglimitsmathematical}
Shao, Z.; Wang, P.; Zhu, Q.; Xu, R.; Song, J.; Bi, X.; Zhang, H.; Zhang, M.; Li, Y.~K.; Wu, Y.; and Guo, D. 2024.
\newblock DeepSeekMath: Pushing the Limits of Mathematical Reasoning in Open Language Models.
\newblock arXiv:2402.03300.

\bibitem[{Su et~al.(2025)Su, Kong, Lin, Jennings, Norick, Kliegl, Patwary, Shoeybi, and Catanzaro}]{su2024nemotron}
Su, D.; Kong, K.; Lin, Y.; Jennings, J.; Norick, B.; Kliegl, M.; Patwary, M.; Shoeybi, M.; and Catanzaro, B. 2025.
\newblock Nemotron-{CC}: Transforming {C}ommon {C}rawl into a Refined Long-Horizon Pretraining Dataset.
\newblock In Che, W.; Nabende, J.; Shutova, E.; and Pilehvar, M.~T., eds., \emph{Proceedings of the 63rd Annual Meeting of the Association for Computational Linguistics (Volume 1: Long Papers)}, 2459--2475. Vienna, Austria: Association for Computational Linguistics.
\newblock ISBN 979-8-89176-251-0.

\bibitem[{Suzgun et~al.(2023)Suzgun, Scales, Sch{\"a}rli, Gehrmann, Tay, Chung, Chowdhery, Le, Chi, Zhou, and Wei}]{suzgun-etal-2023-challenging}
Suzgun, M.; Scales, N.; Sch{\"a}rli, N.; Gehrmann, S.; Tay, Y.; Chung, H.~W.; Chowdhery, A.; Le, Q.; Chi, E.; Zhou, D.; and Wei, J. 2023.
\newblock Challenging {BIG}-Bench Tasks and Whether Chain-of-Thought Can Solve Them.
\newblock In \emph{Findings of the Association for Computational Linguistics: ACL 2023}, 13003--13051.

\bibitem[{Tong et~al.(2024)Tong, Zhang, Wang, Wu, and He}]{tong2024dartmathdifficultyawarerejectiontuning}
Tong, Y.; Zhang, X.; Wang, R.; Wu, R.; and He, J. 2024.
\newblock {DART}-Math: Difficulty-Aware Rejection Tuning for Mathematical Problem-Solving.
\newblock In \emph{The Thirty-eighth Annual Conference on Neural Information Processing Systems}.

\bibitem[{Villalobos et~al.(2024)Villalobos, Ho, Sevilla, Besiroglu, Heim, and Hobbhahn}]{villalobos2024rundatalimitsllm}
Villalobos, P.; Ho, A.; Sevilla, J.; Besiroglu, T.; Heim, L.; and Hobbhahn, M. 2024.
\newblock Will we run out of data? Limits of LLM scaling based on human-generated data.
\newblock arXiv:2211.04325.

\bibitem[{Wang et~al.(2024)Wang, Ma, Zhang, Ni, Chandra, Guo, Ren, Arulraj, He, Jiang, Li, Ku, Wang, Zhuang, Fan, Yue, and Chen}]{wang2024mmlupro}
Wang, Y.; Ma, X.; Zhang, G.; Ni, Y.; Chandra, A.; Guo, S.; Ren, W.; Arulraj, A.; He, X.; Jiang, Z.; Li, T.; Ku, M.; Wang, K.; Zhuang, A.; Fan, R.; Yue, X.; and Chen, W. 2024.
\newblock {MMLU}-Pro: A More Robust and Challenging Multi-Task Language Understanding Benchmark.
\newblock In \emph{The Thirty-eight Conference on Neural Information Processing Systems Datasets and Benchmarks Track}.

\bibitem[{Wang et~al.(2025)Wang, Zhou, Li, and Liu}]{wang2025octothinkermidtrainingincentivizesreinforcement}
Wang, Z.; Zhou, F.; Li, X.; and Liu, P. 2025.
\newblock OctoThinker: Mid-training Incentivizes Reinforcement Learning Scaling.
\newblock arXiv:2506.20512.

\bibitem[{Wettig et~al.(2024)Wettig, Gupta, Malik, and Chen}]{wettig2024qurating}
Wettig, A.; Gupta, A.; Malik, S.; and Chen, D. 2024.
\newblock QuRating: Selecting High-Quality Data for Training Language Models.
\newblock In \emph{International Conference on Machine Learning}, 52915--52971.

\bibitem[{Yang et~al.(2025)Yang, Band, Li, Candes, and Hashimoto}]{yang2024syntheticcontinuedpretraining}
Yang, Z.; Band, N.; Li, S.; Candes, E.; and Hashimoto, T. 2025.
\newblock Synthetic continued pretraining.
\newblock In \emph{The Thirteenth International Conference on Learning Representations}.

\bibitem[{Zellers et~al.(2019)Zellers, Holtzman, Bisk, Farhadi, and Choi}]{zellers-etal-2019-hellaswag}
Zellers, R.; Holtzman, A.; Bisk, Y.; Farhadi, A.; and Choi, Y. 2019.
\newblock {H}ella{S}wag: Can a Machine Really Finish Your Sentence?
\newblock In \emph{Proceedings of the 57th Annual Meeting of the Association for Computational Linguistics}, 4791--4800.

\bibitem[{Zhou et~al.(2025)Zhou, Wang, Ranjan, Cheng, Tang, He, Liu, and Xing}]{zhou2025megamath}
Zhou, F.; Wang, Z.; Ranjan, N.; Cheng, Z.; Tang, L.; He, G.; Liu, Z.; and Xing, E.~P. 2025.
\newblock MegaMath: Pushing the Limits of Open Math Corpora.
\newblock arXiv:2504.02807.

\bibitem[{Zhou et~al.(2024)Zhou, Zhang, jiapeng wang, Chen, Zhao, Sha, Sheng, Wang, and Wen}]{zhou2024jiuzhang}
Zhou, K.; Zhang, B.; jiapeng wang; Chen, Z.; Zhao, X.; Sha, J.; Sheng, Z.; Wang, S.; and Wen, J.-R. 2024.
\newblock JiuZhang3.0: Efficiently Improving Mathematical Reasoning by Training Small Data Synthesis Models.
\newblock In \emph{The Thirty-eighth Annual Conference on Neural Information Processing Systems}.

\end{thebibliography}

\appendix
\renewcommand{\thefigure}{A\arabic{figure}} 
\renewcommand{\thetable}{A\arabic{table}} 
\setcounter{figure}{0}
\setcounter{table}{0}

\section{Experimental Details \label{sec:appendix_train}}
\subsection{Training Details \label{sec:appendix_train_details}}
We employ DeepSeek-R1 for data synthesis and DeepSeek-V3 for answer refinement, with both models setting temperature to 0.6, top-p value to 0.95, and top-k to -1. All computations are executed on a dedicated cluster of 300 H20-141G GPUs.

We use 256 Ascend 910B NPUs to continually pre-train the Llama-3 8B model from the 2T-token checkpoint using 40B tokens of QA blend with KnowEdu, each model taking over 22 hours. We implement it via the Megatron framework~\citep{shoeybi2020megatronlmtrainingmultibillionparameter}, optimized by the Adam algorithm with standard $\beta_1=0.9$ and $\beta_2=0.95$ parameters. The training employs a global batch size of 960 and a linearly decaying learning rate schedule initialized at $1.9\times 10^{-4}$ and terminating at $1.9\times 10^{-5}$.

In the model size scale experiment, we also test the performance of the dataset on the 1.7B and 16B models with the same settings as 8B. For the 1.7B model, we use 80 NPUs for training, each model taking over 38 hours. For the 16B model, we use 480 NPUs for training, each model taking over 21 hours. In Table~\ref{tab:appendix_train_hyper}, we present the model configuration of the 1.7B and 8B models.

We further analyze the computational cost and data scale for constructing 1M QA samples. Specifically, generating 1M QA pairs using DeepSeek-R1 requires 514.07 GPU hours on H20 GPUs, while answer refinement with DeepSeek-V3 costs an additional 318.72 GPU hours. For reference, 1M pure QA samples correspond to 0.093B tokens, and 1M CoT-augmented QA samples correspond to 0.437B tokens.
\begin{table}[htbp]
    \centering
    \begin{tabular}{l|c|c|c}
        \toprule
        \small
        \textbf{Hyperparameter}	& \textbf{1.7B} & \textbf{8B} & \textbf{16B} \\
        \midrule
        Precision	& bfloat16 & bfloat16 & bfloat16 \\
        Layers	& 24 & 32 & 40 \\
        Hidden Size	& 2048 & 4096 & 5120 \\
        Attention Heads & 32 & 32 & 64 \\
        Head Type	& GQA & GQA & GQA \\
        Intermediate Size	& 8192 & 14336 & 18432 \\
        Vocab Size	& 131072 & 131072 & 163840 \\
        Sequence Length	& 8192 & 8192 & 8192 \\
        Activation	& SiLU & SiLU & SiLU \\
        Position Embedding	& RoPE & RoPE & RoPE \\
        \bottomrule
    \end{tabular}
    \caption{Model structure of Llama-3 1.7B, 8B, and 16B.}
    \label{tab:appendix_train_hyper}
\end{table}

\subsection{Datasets \label{sec:appendix_datasets}}

\begin{table}[ht]
    \centering
    \setlength{\tabcolsep}{1mm}
    \small
    \begin{tabular}{l|ccccc}
        \toprule
        \textbf{Dataset} & \textbf{Synthesis}	& \textbf{CoT} &\textbf{Type}	&\textbf{Domain}	&\textbf{Tokens} \\
        \midrule
        N-CC		&\CheckmarkBold &\XSolidBrush & Doc. + QA	& General	& 499.5B \\
        N-CC QA &\CheckmarkBold &\XSolidBrush & QA & General & 51B \\
        YulanQA		&\XSolidBrush &\CheckmarkBold & QA	& General	& 4.92B \\ 
        N-MIND		&\CheckmarkBold &\CheckmarkBold & Conv.	& Math	& 138B \\
        MegaMathQA		&\CheckmarkBold &\CheckmarkBold & QA	& Math	& 7.0B \\
        JiuZhang3.0		&\CheckmarkBold &\CheckmarkBold & QA	& Math	& 4.6B \\
        \midrule
        LinkQA		&\CheckmarkBold &\XSolidBrush & QA	& General	& 30B \\
        LinkQA$_\text{CoT}$ &\CheckmarkBold &\CheckmarkBold & QA & General & 50B \\
        \bottomrule
    \end{tabular}
    \caption{Comparison with large-scale QA datasets. LinkQA$_\text{CoT}$ extends LinkQA by incorporating supplementary math CoT data LinkQA$_\text{MathCoT}$. In practice, the complete CoT dataset can contain significantly more tokens.}
    \label{tab:appendix_datasets}
\end{table}

We compare LinkQA with large-scale QA datasets of different types and from different sources, detailed in Table~\ref{tab:appendix_datasets}. For Nemotron-CC (N-CC), we maintain its original 9:1 document-to-QA ratio during experiments. We decompose it into Nemotron-CC Document and Nemotron-CC QA components. To isolate document effects, we substitute Nemotron-CC Document with KnowEdu as an alternative experimental setting and report optimal configurations as the result of N-CC.

\subsection{Scaling Details \label{sec:appendix_scaling}}
The specific accuracy values of the final checkpoint in the scaling experiments are shown in Table~\ref{tab:appendix_scaling}.

\begin{table}[ht]
    \centering
    \setlength{\tabcolsep}{1mm}
    \small
    \begin{tabular}{l|l|cccc}
        \toprule
        \textbf{Scale} & \textbf{Settings} & \textbf{MMLU} & \textbf{CMMLU} & \textbf{C-Eval} & \textbf{STEM} \\
        \midrule
        \textbf{Model} & 1.7B: KnowEdu & 45.32 & 48.95 & 45.39 & 38.63 \\     
        \textbf{Size} & 1.7B: LinkQA & 52.56 & 54.00 & 53.63 & 44.85 \\
        & 8B: KnowEdu & 63.53 & 68.08 & 67.18 & 53.86 \\
        & 8B: LinkQA & 68.57 & 70.89 & 70.15 & 60.91 \\
        & 16B: KnowEdu & 68.61 & 72.53 & 69.23 & 61.44 \\
        & 16B: LinkQA & 70.45 & 75.03 & 73.30 & 64.91 \\
        \midrule
        & 2T: KnowEdu & 58.17 & 62.99 & 61.98 & 49.16 \\
        \textbf{Initial} & 2T: LinkQA & 64.40 & 66.35 & 65.59 & 56.95 \\
        \textbf{FLOPs} & 10T: KnowEdu & 63.53 & 68.08 & 67.18 & 53.86 \\
        & 10T: LinkQA & 68.57 & 70.89 & 70.15 & 60.91 \\
        \bottomrule
    \end{tabular}
    \caption{Accuracy of the final checkpoint in the scaling experiments. Abbreviations: STEM = MMLU-STEM. For model size scalability, we use initial checkpoints of 4T, 10T, and 10T tokens for 1.7B, 8B, and 16B parameter models, respectively.}
    \label{tab:appendix_scaling}
\end{table}
\subsection{Seed Data Distribution\label{sec:Seed_Data_Distribution}}
\begin{figure*}[t]
    \centering
    \includegraphics[width=0.95\textwidth]{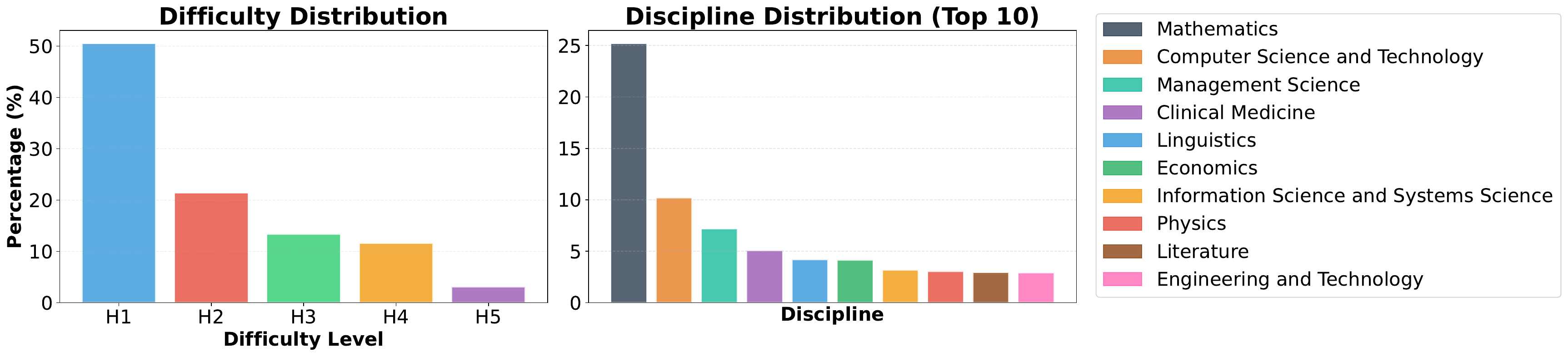}
    \caption{Distribution of all seed data: difficulty distribution (left) and discipline distribution (right).}
    \label{fig:all_seed_data_distribution}
\end{figure*}
\begin{figure*}[t]
    \centering
    \includegraphics[width=0.95\textwidth]{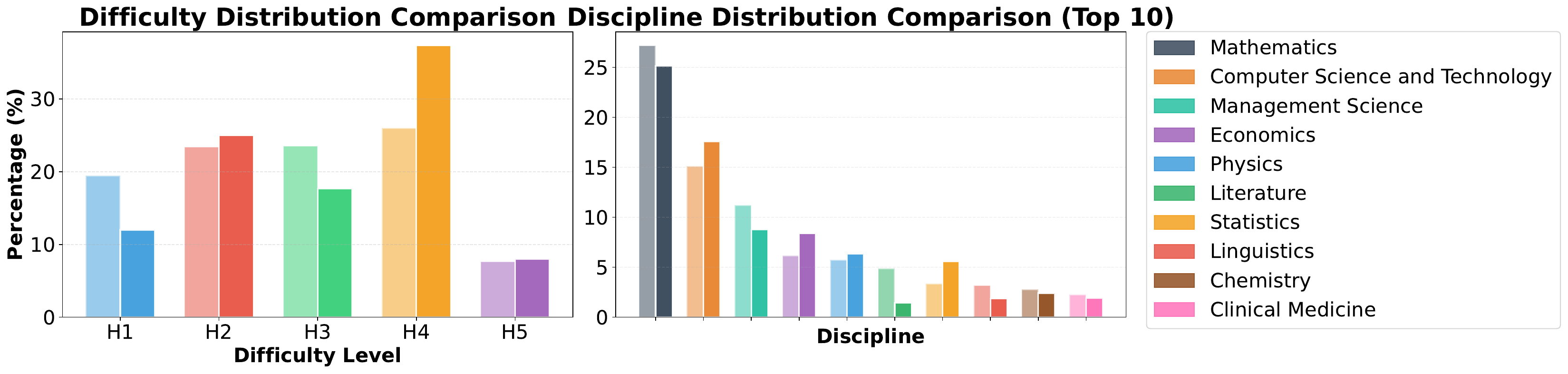}
    \caption{Distribution comparison of LinkSyn sampled seed data (light color) and LinkQA (dark color): difficulty distribution (left) and discipline distribution (right).}
    \label{fig:seed_gen_data_distribution}
\end{figure*}
The difficulty and discipline distribution of our seed data are illustrated in the Figure~\ref{fig:all_seed_data_distribution}. Regarding discipline distribution, the seed data covers all first-level disciplines with balanced proportions, where mathematics accounts for the largest share at 25\% but remains within reasonable bounds. However, the difficulty distribution shows significant imbalance, with 50\% of the seed data concentrated at the H1 difficulty level. To address this imbalance, we implement difficulty control measures during the data sampling process.

\subsection{LinkQA Distribution\label{sec:LinkQA_distribution}}
The difficulty and subject distributions of our sampled seed data and LinkQA (we sampled 100k data for difficulty and subject annotation analysis) are shown in Figure~\ref{fig:seed_gen_data_distribution}. Our target difficulty distribution is (H1–H5, from easiest to hardest): 10\%, 15\%, 25\%, 25\%, 25\%, but due to the scarcity of high-difficulty data, the actual difficulty can only approximate the target difficulty as described in Equation~\ref{eq:instance-selection}, resulting in some deviation in our final sampled seed data. Notably, LinkQA exhibits a higher proportion of high-difficulty questions compared to the seed data because multi-seed data synthesis tends to elevate the overall difficulty level. Regarding subject distribution, both datasets approximate natural distributions, though we observe that LinkQA shows reduced proportions in general subjects such as Mathematics and Literature compared to the sampled seed data. This occurs because knowledge points in these general subjects have more connections with other disciplines, leading to cross-disciplinary data appearing in the sampled knowledge point paths, which causes the subject distribution inconsistency between LinkQA and seed data.

\subsection{Quality Review of LinkQA\label{sec:appendix_qualityreview}}

To rigorously assess the quality of LinkQA, we randomly sample 100 QA pairs from each predefined difficulty level. A professional annotator evaluates each pair along two dimensions: (i) the solvability of the question, and (ii) the accuracy of the corresponding answer. A QA pair is deemed correct only if the question is solvable and the provided answer is fully accurate. The results, summarized in Table~\ref{tab:qualityreview}, demonstrate that the majority of synthetic QA pairs meet the correctness criterion. Moreover, the correctness rate exhibits a decreasing trend with increasing difficulty, indicating a correlation between difficulty level and quality metrics.

\begin{table}[ht]
    \centering
    \begin{tabular}{lc}
        \toprule
        \textbf{Difficulty Level} & \textbf{Correct (\%)} \\
        \midrule
        H1/H2 & 98 \\
        H3    & 94 \\
        H4/H5 & 87 \\
        \bottomrule
    \end{tabular}
    \caption{Manual quality review results for LinkQA across different difficulty levels.}
    \label{tab:qualityreview}
\end{table}

\section{Knowledge Point Graph}
\subsection{Knowledge Point Consolidation\label{app:dedup}}
\begin{figure}[htbp]
    \centering
    \includegraphics[width=0.48\textwidth]{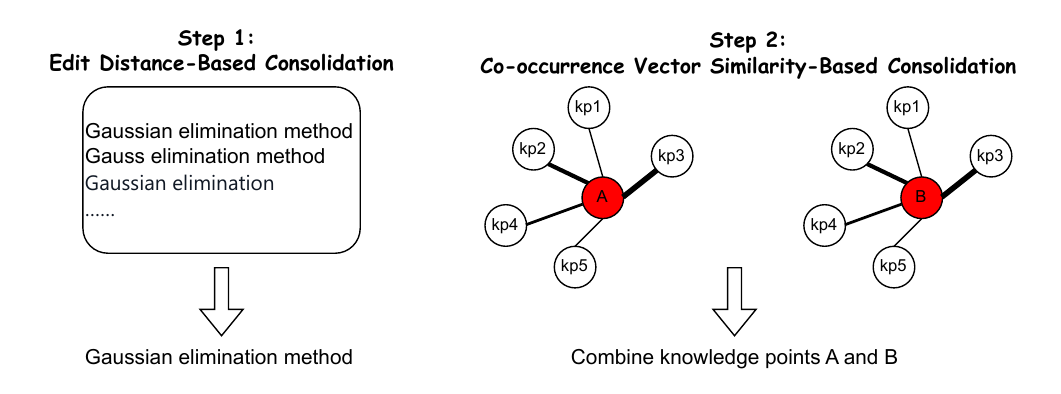}
    \caption{Two-step knowledge point consolidation process: edit distance-based deduplication (Step 1) and co-occurrence vector similarity-based deduplication (Step 2).}
    \label{fig:kp_combination}
\end{figure}

As illustrated in Figure~\ref{fig:kp_combination}, we perform knowledge point (KP) consolidation in two stages. In the first stage, we standardize the case of all KPs, then group KPs by the first three identical characters (prefix length 3), and within each group, we cluster KPs such that the maximum pairwise edit distance in a cluster does not exceed the greater of 3 or $\mathrm{int}(0.5 \times \max(\mathrm{len}(s_1), \mathrm{len}(s_2)))$. For each cluster, we use Qwen-14B to summarize and merge the KPs. In the second stage, we compute co-occurrence vectors for each KP and cluster those with cosine similarity above 0.9; again, we use Qwen-14B to summarize and consolidate each cluster. This detailed consolidation process improves the KP graph, but as it does not critically affect LinkSyn, therefore, the consolidation parameters are flexible and can be adjusted as needed. In our implementation, these steps result in a final set of 10M KPs.
\begin{figure}[htbp]
    \centering
    \includegraphics[width=0.9\columnwidth]{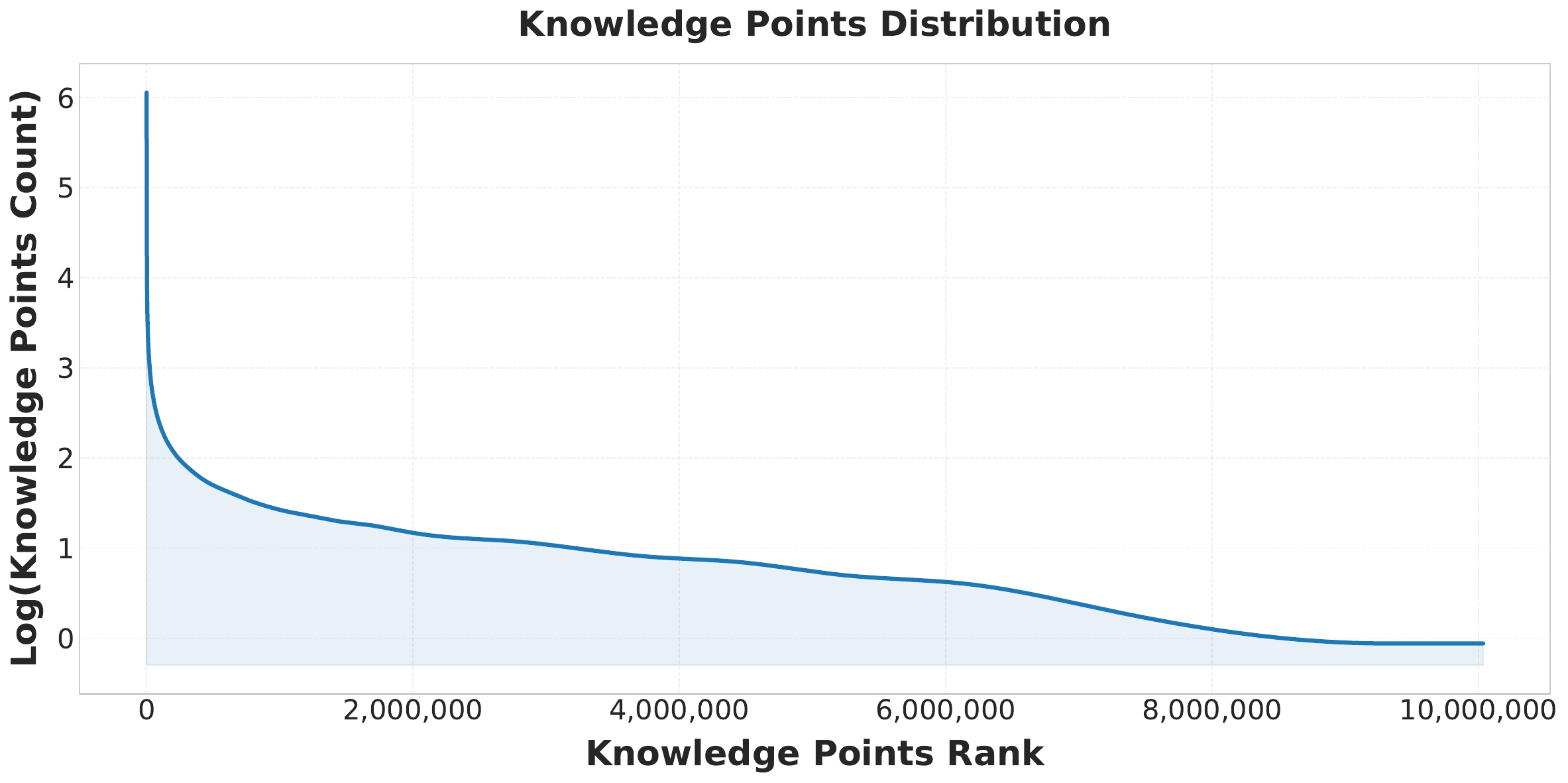}
    \caption{Knowledge points frequency distribution.}
    \label{fig:distributions}
\end{figure}
\subsection{Knowledge Point Graph Analysis\label{sec:kp_graph_analysis}}
The knowledge point graph encompasses 10M nodes interconnected by 153M edges, exhibiting a notably sparse graph structure. As illustrated in Figure~\ref{fig:distributions}, both the text quantity distribution across nodes (ranging from 1 to 737,794) and edge weight distribution (ranging from 1 to 149,382) exhibit pronounced coverage characteristics. We performed connectivity analysis on the graph, revealing that our knowledge point network constructed from seed data consists of one giant connected component containing over 92\% of texts and more than 89\% of knowledge points (with a diameter of 29), alongside numerous smaller connected components. These smaller components contain fewer than 44 knowledge points each and are consistently confined to single discipline domains. The network's assortativity coefficient is merely 0.0892, indicating limited degree homophily. We apply the Leiden community detection algorithm on the graph and get 21,806 knowledge point clusters, with the dominant subject in each cluster averaging 86.76\% of content, demonstrating that knowledge points naturally aggregate according to their disciplinary boundaries while maintaining crucial cross-domain connections.

\onecolumn
\subsection{Uniform Distribution Maximizes Coverage\label{app:coverage_uniform_proof}}
Let $K = \{k_1, k_2, \ldots, k_n\}$ denote the set of knowledge points, and let $p = (p_1, p_2, \ldots, p_n)$ be a sampling distribution on the $(n-1)$-dimensional probability simplex, i.e., $p_i \geq 0$ and $\sum_{i=1}^n p_i = 1$. Suppose we perform $M$ independent samples in total. For each knowledge point $k_i$, let $N(k_i)$ denote the number of times $k_i$ is sampled. The probability that $k_i$ is not sampled in any of the $M$ draws is $(1-p_i)^M$, so the probability that it is sampled at least once is $1-(1-p_i)^M$. The expected number of distinct knowledge points sampled, or the expected coverage, is therefore
$$
C(p) = \mathbb{E}\left[\sum_{i=1}^n \mathbb{I}(N(k_i)>0)\right] = \sum_{i=1}^n \left[1 - (1-p_i)^M\right],
$$
where $\mathbb{I}(\cdot)$ is the indicator function.

To maximize $C(p)$ over all valid probability distributions $p$, we observe that each term $1-(1-p_i)^M$ is strictly concave in $p_i$ for $p_i \in (0,1)$ and $M \geq 1$, as the second derivative satisfies $-M(M-1)(1-p_i)^{M-2} < 0$. Thus, $C(p)$ is a strictly concave function on the probability simplex and is also symmetric with respect to all $p_i$. By symmetry and concavity, the maximum of $C(p)$ is achieved when all $p_i$ are equal. Imposing the constraint $\sum_{i=1}^n p_i = 1$ yields the unique solution $p_i^* = \frac{1}{n}$ for all $i$. Therefore, the uniform distribution $p^a_i = \frac{1}{n}$ for all $k_i \in K$ uniquely maximizes the expected coverage $C(p)$.

\subsection{Convexity Properties of the Optimal Knowledge Distribution\label{app:kv_optimal_proof}}
We demonstrate that the optimal sampling policy $p^*$ minimizing the Knowledge Value function $\mathrm{KV}(p)$ is a convex combination of the uniform distribution $p^a$ and the empirical distribution $p^b$. The proof covers both squared Euclidean distance and reverse KL divergence as divergence measures.

\paragraph{Case 1: Squared Euclidean Distance}

When using squared Euclidean distance as the divergence measure, the Knowledge Value function becomes:
$$
\mathrm{KV}(p) = \beta\|p - p^a\|_2^2 + (1-\beta)\|p - p^b\|_2^2,
$$
where $\beta \in [0,1]$, $p^a$ is the uniform distribution, and $p^b$ is the empirical distribution. Our goal is to find the distribution $p^*$ that minimizes $\mathrm{KV}(p)$ subject to the constraints that $p$ is a probability distribution, i.e., $p_i \geq 0$ for all $i$ and $\sum_{i=1}^n p_i = 1$.

Expanding the squared Euclidean distances, we have:
$$
\mathrm{KV}(p) = \beta\sum_{i=1}^n (p_i - p^a_i)^2 + (1-\beta)\sum_{i=1}^n (p_i - p^b_i)^2.
$$

Further expansion yields:
$$
\mathrm{KV}(p) = \sum_{i=1}^n \left[\beta(p_i^2 - 2p_i p^a_i + (p^a_i)^2) + (1-\beta)(p_i^2 - 2p_i p^b_i + (p^b_i)^2)\right].
$$

Rearranging terms:
$$
\mathrm{KV}(p) = \sum_{i=1}^n \left[p_i^2 - 2p_i(\beta p^a_i + (1-\beta)p^b_i)\right] + C,
$$
where $C$ is a constant independent of $p$.

To minimize $\mathrm{KV}(p)$ subject to the constraints, we form the Lagrangian:
$$
L(p, \lambda) = \sum_{i=1}^n \left[p_i^2 - 2p_i(\beta p^a_i + (1-\beta)p^b_i)\right] + \lambda\left(\sum_{i=1}^n p_i - 1\right).
$$

Taking the partial derivative with respect to each $p_i$ and setting it to zero:
$$
\frac{\partial L}{\partial p_i} = 2p_i - 2(\beta p^a_i + (1-\beta)p^b_i) + \lambda = 0.
$$

Solving for $p_i$:
$$
p_i = \beta p^a_i + (1-\beta)p^b_i - \frac{\lambda}{2}.
$$

Using the constraint $\sum_{i=1}^n p_i = 1$:
$$
\sum_{i=1}^n p_i = \sum_{i=1}^n \left[\beta p^a_i + (1-\beta)p^b_i - \frac{\lambda}{2}\right] = \beta\sum_{i=1}^n p^a_i + (1-\beta)\sum_{i=1}^n p^b_i - \frac{n\lambda}{2} = 1.
$$

Since $p^a$ and $p^b$ are probability distributions, $\sum_{i=1}^n p^a_i = \sum_{i=1}^n p^b_i = 1$, so:
$$
\beta \cdot 1 + (1-\beta) \cdot 1 - \frac{n\lambda}{2} = 1 \implies \frac{n\lambda}{2} = 0 \implies \lambda = 0.
$$

Therefore, the optimal solution is:
$$
p_i^* = \beta p^a_i + (1-\beta)p^b_i.
$$

This is a valid probability distribution because $p^a$ and $p^b$ are probability distributions and $\beta \in [0,1]$, so $p_i^* \geq 0$ for all $i$ and $\sum_{i=1}^n p_i^* = 1$. The Hessian matrix of the objective function is $2I$, which is positive definite, confirming that $p^*$ is the unique global minimizer.

\paragraph{Case 2: Reverse KL Divergence}

When using reverse KL divergence, the Knowledge Value function is:
$$
\mathrm{KV}(p) = \beta\,\mathrm{KL}(p^a\|p) + (1-\beta)\,\mathrm{KL}(p^b\|p),
$$
where the reverse KL divergence is defined as:
$$
\mathrm{KL}(q\|p) = \sum_{i=1}^n q_i\log\frac{q_i}{p_i}.
$$

Expanding $\mathrm{KV}(p)$:
$$
\mathrm{KV}(p) = \beta\sum_{i=1}^n p^a_i\log\frac{p^a_i}{p_i} + (1-\beta)\sum_{i=1}^n p^b_i\log\frac{p^b_i}{p_i}.
$$

Further expansion gives:
$$
\mathrm{KV}(p) = \beta\sum_{i=1}^n p^a_i\log p^a_i + (1-\beta)\sum_{i=1}^n p^b_i\log p^b_i - \sum_{i=1}^n\left[\beta p^a_i + (1-\beta)p^b_i\right]\log p_i.
$$

The first two terms are constants with respect to $p$, so minimizing $\mathrm{KV}(p)$ is equivalent to maximizing:
$$
\sum_{i=1}^n c_i\log p_i, \quad \text{where } c_i = \beta p^a_i + (1-\beta)p^b_i,
$$
subject to the constraints $p_i \geq 0$ for all $i$ and $\sum_{i=1}^n p_i = 1$.

To solve this constrained optimization problem, we form the Lagrangian:
$$
L(p,\lambda) = \sum_{i=1}^n c_i\log p_i + \lambda\left(1 - \sum_{i=1}^n p_i\right).
$$

Taking the partial derivative with respect to $p_i$ and setting it to zero:
$$
\frac{\partial L}{\partial p_i} = \frac{c_i}{p_i} - \lambda = 0 \implies p_i = \frac{c_i}{\lambda}.
$$

Using the constraint $\sum_{i=1}^n p_i = 1$:
$$
\sum_{i=1}^n p_i = \sum_{i=1}^n \frac{c_i}{\lambda} = \frac{1}{\lambda}\sum_{i=1}^n c_i = 1.
$$

Therefore:
$$
\lambda = \sum_{i=1}^n c_i = \sum_{i=1}^n\left[\beta p^a_i + (1-\beta)p^b_i\right] = \beta\sum_{i=1}^n p^a_i + (1-\beta)\sum_{i=1}^n p^b_i = \beta \cdot 1 + (1-\beta) \cdot 1 = 1.
$$

Thus, the optimal solution is:
$$
p_i^* = c_i = \beta p^a_i + (1-\beta)p^b_i.
$$

This is a valid probability distribution for the same reasons as in the squared Euclidean case. The objective function $\sum_{i=1}^n c_i\log p_i$ is strictly concave in $p$ (as its Hessian has diagonal entries $-\frac{c_i}{p_i^2} < 0$), ensuring that $p^*$ is the unique global maximizer.

In both cases, we have proven that the optimal sampling policy minimizing the Knowledge Value function is $p^* = \beta p^a + (1-\beta)p^b$, which is a convex combination of the uniform distribution and the empirical distribution.

\section{Prompts \label{sec:appendix_prompts}}
\subsection{Data Annotation \label{sec:appendix_prompts_label}}
To efficiently annotate the discipline labels while maintaining quality, we implement a two-stage annotation pipeline using the discipline classifier that categorizes content into 62 first-level disciplines\footnote{GB/T 13745-2008 taxonomy}. Initially, we employ DeepSeek-R1 with discipline-constrained prompts to generate preliminary labels for 20M seed samples. Subsequently, we curate a balanced subset of 500K high-confidence samples through uniform stratified sampling across all 62 disciplines, which is then used to finetune Qwen2.5-7B-Instruct, yielding our specialized subject classifier. Empirical validation demonstrates 82.18\% label consistency between our specialized classifier and DeepSeek-R1, confirming reliable knowledge distillation. The prompt used to annotate data with discipline and train the corresponding labeler is shown as follows.
\begin{tcolorbox}[colback=white!95!gray,colframe=gray!50!black,rounded corners,label={prompt-discipline-classifier}, title={Prompt for Discipline Classifier}]
\begin{lstlisting}[breaklines=true, xleftmargin=0pt, breakindent=0pt, columns=fullflexible, mathescape, numbers=none]
Act as an educational taxonomist. Classify the input question into our standardized discipline hierarchy using sequential reasoning, then output strictly in JSON format:
1. Primary Discipline Identification
   Select exactly one primary discipline from:  
   {Discipline List}  
   - Use "cross-discipline" only for explicit multi-domain integration  
   - Assign "Other" only if no discipline matches >=60% relevance  
2. Secondary Discipline Assignment 
   - Identify the most specific applicable sub-discipline 
   - Null if primary discipline has no sub-domains  
   - Use "General" for non-specialized content  
3. Validation Rules 
   - Reject non-educational content -> Output "Invalid"  
   - Correct spelling/terminology variations before classification  
Output Schema: 
{
  "primary_discipline": "",
  "secondary_discipline": "",
  "confidence": 0.0-1.0,
  "rejection_reason": null
}
Input: {Seed Data}
\end{lstlisting}
\end{tcolorbox}

The list of 62 primary disciplines is as follows:

\begin{tcolorbox}[colback=white!95!gray,colframe=gray!50!black,rounded corners,label={prompt-discipline-list}, title={Discipline List (62)}]
\begin{lstlisting}[breaklines=true, xleftmargin=0pt, breakindent=0pt, columns=fullflexible, mathescape, numbers=none]
['Mathematics', 'Computer Science and Technology', 'Clinical Medicine', 'Chemistry', 'Economics', 'Information Science and Systems Science', 'Physics', 'Biology', 'Law', 'Philosophy', 'Sociology', 'Literature', 'Psychology', 'Statistics', 'History', 'Power and Electrical Engineering', 'Earth Science', 'Management Science', 'Electronics and Communication Technology', 'Linguistics', 'Preventive Medicine and Public Health', 'Political Science', 'Education Science', 'Aerospace Science and Technology', 'Astronomy', 'Materials Science', 'Mechanics', 'Sports Science', 'Ethnology and Cultural Studies', 'Basic Medicine', 'Environmental Science and Resource Science', 'Journalism and Communication', 'Religious Studies', 'Engineering and Technology Related to Information and Systems Science', 'Food Science and Technology', 'Engineering and Technology', 'Art Studies', 'Mechanical Engineering', 'Traditional Chinese Medicine and Chinese Materia Medica', 'Pharmacy', 'Civil and Architectural Engineering', 'Chemical Engineering', 'Nuclear Science and Technology', 'Marxism', 'Agronomy', 'Energy Science and Technology', 'Transportation Engineering', 'Military Science', 'Safety Science and Technology', 'Animal Husbandry and Veterinary Science', 'Archaeology', 'Engineering and Technology Related to Product Applications', 'Library, Information and Documentation Science', 'Geomatics Science and Technology', 'Aquaculture Science', 'Metallurgical Engineering Technology', 'Hydraulic Engineering', 'Military Medicine and Special Medicine', 'Textile Science and Technology', 'Mining Engineering Technology', 'Forestry', 'Engineering and Technology Related to Natural Sciences']
\end{lstlisting}
\end{tcolorbox}
The difficulty scorer operationalizes human performance metrics by defining five difficulty tiers (H1-H5) based on pass rates under standardized one-hour testing conditions with QS Top 100 university students majoring in relevant disciplines. We implement a multi-stage annotation pipeline where initial difficulty annotations are generated by DeepSeek-R1 through structured prompts that simulate human problem-solving behaviors, producing preliminary difficulty estimates for 500K QA pairs. Subsequently, we distill this knowledge by training Qwen2.5-14B-Instruct on the annotated data to create our specialized difficulty classifier. Expert assessment validation with five PhD evaluators (Krippendorff's $\alpha$ = 0.85) confirms strong correlation (Pearson's $r$ = 0.92) between model predictions and actual human performance metrics. The prompt used to annotate data with difficulty labels and train the corresponding labeler is shown as follows.
\begin{tcolorbox}[colback=white!95!gray,colframe=gray!50!black,rounded corners,label={prompt-difficulty-scorer}, title={Prompt for Difficulty Scorer}, breakable]
\begin{lstlisting}[breaklines=true, xleftmargin=0pt, breakindent=0pt, columns=fullflexible, mathescape, numbers=none]
Act as an educational assessment expert, analyze the provided question through sequential reasoning and output strictly in JSON format:  
1. Knowledge Analysis
   - Core concepts (<=3): [comma-separated list]  
   - Integration type: {single-concept | cross-chapter | cross-discipline}  
2. Cognitive Tier (Bloom's Taxonomy)  
   {memory | understanding | application | analysis | synthesis | evaluation}  
3. Difficulty Assessment
   - Estimated pass rate (P) for QS Top 100 university majors: [0-100%]  
   - Tier:  
     - extreme: P < 10%  
     - challenge: 10% <= P < 30%  
     - improvement: 30% <= P < 50%  
     - standard: 50% <= P < 80%  
     - basic: P >= 80%  
     - other: invalid inputs  
4. Exception Handling
   - Mark "other" for non-questions/unanswerable items  
   - Correct minor errors (e.g., missing correct options) before assessment  
   - Ignore provided solutions/answers 

Output Schema:  
{
  "difficulty_tier": "basic|standard|improvement|challenge|extreme|other",
  "rationale": [
    "Involves {N} core knowledge points",
    "Cognitive level: {Bloom's tier}",
    "Estimated pass rate: approximately {XX}% for target cohort"
  ]
}
Input: {Seed Data}
\end{lstlisting}
\end{tcolorbox}
The knowledge point annotator identifies core knowledge concepts within educational content by extracting the most basic and smallest content units that constitute a knowledge system within specific discipline areas. We implement a multi-stage annotation pipeline where DeepSeek-R1 initially generates knowledge point labels for 20M seed samples using structured prompts that employ a hierarchical reasoning approach: first determining discipline classification and educational level assessment, then leveraging this contextual information to more accurately identify up to three core knowledge points per item. The annotation process employs educational taxonomist reasoning that analyzes content while ignoring potentially incorrect solutions, ensuring focus on authentic knowledge concepts. Subsequently, we curate a balanced training dataset through stratified sampling across multiple disciplines from the annotated seed data. This curated dataset is then used to finetune Qwen2.5-14B-Instruct, yielding our specialized knowledge point classifier that achieves an 80.08\% agreement rate with DeepSeek-R1’s annotations when allowing an edit distance of up to 3 on held-out test data. The prompt used to annotate data with knowledge point labels and train the corresponding classifier is shown as follows.

\begin{tcolorbox}[colback=white!95!gray,colframe=gray!50!black,rounded corners,label={prompt-kp-extractor}, title={Prompt for Knowledge Point Annotation}, breakable]
\begin{lstlisting}[breaklines=true, xleftmargin=0pt, breakindent=0pt, columns=fullflexible, mathescape, numbers=none]
Act as an educational taxonomist. Analyze the provided item through step-by-step reasoning and output strictly in JSON format:
1. discipline Classification
   - Identify the discipline to which the item belongs.
   - discipline list: {Discipline List}
2. Educational Level
   - Choose from: [Elementary School, Middle School, High School, University, Graduate School]
3. Knowledge Point Analysis
    - Core knowledge points (<=3): [comma-separated list]
    - Knowledge Point Definition: A knowledge point refers to the most basic and smallest content unit that constitutes a knowledge system within a certain discipline area. 
    - Example:
     - Mathematics: Properties of linear functions
     - English: Present perfect tense
     - Biology: Basic laws of heredity
4. Exception Handling
   - Ignore any provided solutions or answer steps, as they may be incorrect or suboptimal.
   - Only select from the provided candidate lists for discipline, Assessment Ability, and Educational Level.
Output Schema:
{
  "Knowledge Point List": [
    "Properties of linear functions"
    ...
  ]
}
Input: {Seed Data}
\end{lstlisting}
\end{tcolorbox}

\subsection{Synthesis Prompts \label{sec:appendix_prompts_synthesis}}
The prompts and specific rules used to synthesize diverse knowledge-intensive QA pairs are as follows.
\begin{tcolorbox}[colback=white!95!gray,colframe=gray!50!black,rounded corners,label={prompt-multi-grade}, title={Prompt for Synthesizer}, breakable]
\begin{lstlisting}[breaklines=true, xleftmargin=0pt, breakindent=0pt, columns=fullflexible, mathescape, numbers=none]
Act as a {Role Assigner} educator, analyze the knowledge points assessed by the provided {ref_num] reference questions. Generate {gen_num} novel questions adhering to these requirements:
1. Questions must demonstrate substantial differentiation while testing application or higher-order use of identified knowledge points.
2. Difficulty must align with high-difficulty standards through:
   a) Down-scaling overqualified knowledge points to prerequisite concepts at graduate level
   b) Up-scaling underqualified points to advanced applications at graduate level
3. Linguistic consistency must be maintained with the input questions.
[Difficulty Reference Guide]
1. Knowledge Analysis:  
   - Core concepts (<=3)  
   - Integration type: {single | cross-chapter | cross-discipline}  
2. Cognitive Tier (Bloom's Taxonomy):  
   {memory | understanding | application | analysis | synthesis | evaluation}  
3. Difficulty Calibration:  
   - Estimate pass rate 0 <= P <= 100%
   - Tier Classification:  
     - extreme: P < 10%  
     - challenge: 10% <= P < 30%  
     - improvement: 30% <= P < 50%  
     - standard: 50% <= P < 80%  
     - basic: P >= 80%   
   - ENSURE generated questions match reference difficulty tier
   
Output Schema: {Format-specified JSON}
Input: {Seed Data}
\end{lstlisting}
\end{tcolorbox}
\texttt{\{Role Assigner\}} can be set to ``college'' or ``graduate''. In our experiment, \texttt{\{ref\_num\}} represents the number of reference QA pairs, and \texttt{\{gen\_num\}} represents the number of synthetic QA pairs to generate. The typical mapping is: when \texttt{\{ref\_num\}} = 1, \texttt{\{gen\_num\}} = 10; when \texttt{\{ref\_num\}} = 2, \texttt{\{gen\_num\}} = 15; when \texttt{\{ref\_num\}} = 3, \texttt{\{gen\_num\}} = 20.

Here is the prompt for regenerating answers to the generated questions:
\begin{tcolorbox}[colback=white!95!gray,colframe=gray!50!black,rounded corners,label={prompt-ans-refine}, title={Prompt for Answer Regenerator}]
\begin{lstlisting}[breaklines=true, xleftmargin=0pt, breakindent=0pt, columns=fullflexible, mathescape, numbers=none]
Please strictly follow the requirements below to analyze the given question and answer:
Answer Requirements
1. Perform step-by-step reasoning and show the complete thought process, which must include:
   - Extraction of key information from the question
   - Application of relevant formulas/theorems
   - Analysis of each option individually
   - Reminders of common error types
   - Display of logical reasoning chains
2. Answer format requirements:
   - Must include both 'Solution Steps' and 'Final Answer' fields
3. Notes
   - If the question already includes solution steps and answers, please ignore them and don't be influenced by them, as they may be incorrect or suboptimal.
   - For multiple-choice questions:
     * If the correct answer is missing:
         - Add a fifth option: "(E) [Correct Answer]"
         - Set answer_index=4
         - Keep the original options unchanged
{Format-specific Constraints}
Output Schema: {Format-specified JSON}
Input: {Question}
\end{lstlisting}
\end{tcolorbox}

\texttt{\{Format-specific Constraints\}} and \texttt{\{Format-specific JSON\}} are controlled by the rule enforcer and vary depending on whether the targeted synthetic question type is multiple-choice or essay-question format, following the specific rules below:

\begin{tcolorbox}[colback=white!95!gray,colframe=gray!50!black,rounded corners,label={prompt-rules}, title={Rules for Targeted Synthetic Qustion Type}]
\begin{lstlisting}[breaklines=true, xleftmargin=0pt, breakindent=0pt, columns=fullflexible, mathescape, numbers=none]
Format-specific Constraints:
Multiple-Choice: 4. The generated question type is multiple-choice. For each question, four alternative options must be generated, and among the four options, there must be one correct answer.
Essay-question: 4. The generated question type is essay-question. For each question, the solution steps and the final correct answer are provided. The generated questions cannot be open-ended questions (such as those of the solution type, thinking type, information listing type, etc.), but must be self-contained with a final answer that can be determined as correct.

Format-specified JSON:
Multiple-Choice: [{"question": "", "options": [],  "answer_index": 0-3}, ...]
Essay-Question: [{"question": "", "solution": "",  "answer": ""}, ...]

\end{lstlisting}
\end{tcolorbox}

\section{Case Study \label{sec:Case_Study}}

\begin{longtable}{p{0.16\textwidth}|p{0.16\textwidth}|p{0.16\textwidth}|p{0.16\textwidth}|p{0.16\textwidth}|p{0.16\textwidth}}
\caption{Case of different number of seed data and different question types.}
\label{tab:case_seed_syn} \\
\toprule
\midrule
\endfirsthead

\caption[]{(continued)} \\
\toprule
\midrule
\endhead

\midrule
\endfoot

\bottomrule
\endlastfoot

\multicolumn{6}{p{0.96\textwidth}}{\textbf{Diffusion with one seed data (multiple-choice, n = 10)}} \\
\midrule
\multicolumn{6}{p{0.96\textwidth}}{\textbf{Knowledge point path}} \\
\midrule
\multicolumn{6}{p{0.96\textwidth}}{Differential diagnosis of functional vs. organic abdominal pain}
\\
\midrule
\multicolumn{6}{p{0.96\textwidth}|}{\textbf{Seed Data 1}}
\\
\midrule
\multicolumn{6}{p{0.96\textwidth}}{\small \textbf{Question}: I am a healthy 24-year-old female with a lifelong history of IBS. When I stand for a long time, I sometimes feel discomfort or heaviness under my left rib, in the upper abdomen, like there's a brick in there. It's not painful, just uncomfortable, especially after walking or standing for a couple of hours. What could be causing this and what should I do?}\\
\multicolumn{6}{p{0.96\textwidth}}{\small \textbf{Answer}: This discomfort could be related to your IBS, possibly due to gas accumulation or splenic flexure syndrome (trapped gas in the colon bend). Prolonged standing may worsen digestive symptoms or cause muscle strain. Consider: 1) Avoiding trigger foods and managing IBS symptoms, 2) Taking breaks during long periods of standing, 3) Gentle stretching and posture improvement. Consult your doctor if symptoms worsen or new symptoms develop, as proper evaluation may be needed to rule out other causes.}
\\
\midrule
\multicolumn{6}{p{0.96\textwidth}}{\textbf{Synthetic QA pairs}} \\
\midrule
\multicolumn{6}{p{0.96\textwidth}}{\small \textbf{Question 1}: A patient with IBS complains of dull pain in the left upper abdomen after prolonged standing, with no peritoneal irritation on examination. What is the most likely underlying mechanism? 
A. Superior mesenteric artery compression syndrome \quad B. Diaphragmatic reflux caused by increased intra-abdominal pressure \quad C. Visceral hypersensitivity-induced referred pain \quad D. Intercostal nerve entrapment syndrome }
\\
\multicolumn{6}{p{0.96\textwidth}}{\small \textbf{Answer 1}: C. Visceral hypersensitivity-induced referred pain }
\\
\multicolumn{6}{p{0.96\textwidth}}{\small \textbf{Question 2}: What is the key indicator to distinguish IBS-related abdominal pain from early chronic pancreatitis? 
A. Serum amylase fluctuation \quad B. Temporal relationship between pain and defecation \quad C. Abdominal CT findings \quad D. Change in body mass index }
\\
\multicolumn{6}{p{0.96\textwidth}}{\small \textbf{Answer 2}: B. Temporal relationship between pain and defecation }
\\
\multicolumn{6}{p{0.96\textwidth}}{\small \textbf{Question 3}: When standing induces abdominal pain, which test is most specific for the diagnosis of hiatal hernia? 
A. 24-hour esophageal pH monitoring \quad B. Standing abdominal ultrasound \quad C. Enhanced CT 3D reconstruction \quad D. Dynamic barium meal observation }
\\
\multicolumn{6}{p{0.96\textwidth}}{\small \textbf{Answer 3}: D. Dynamic barium meal observation }
\\
\multicolumn{6}{p{0.96\textwidth}}{\small \textbf{Question 4}: Which neural coordination disorder is most likely to cause postural (standing-induced) abdominal pain? 
A. Vagus nerve and greater splanchnic nerve \quad B. Intercostal nerve and iliohypogastric nerve \quad C. Pelvic nerve and pudendal nerve \quad D. Phrenic nerve and subcostal nerve }
\\
\multicolumn{6}{p{0.96\textwidth}}{\small \textbf{Answer 4}: A. Vagus nerve and greater splanchnic nerve }
\\
\multicolumn{6}{p{0.96\textwidth}}{\small \textbf{Question 5}: What hemodynamic parameter should be especially noted when postural change induces abdominal pain? 
A. Portal venous pressure gradient \quad B. Mesenteric vascular resistance \quad C. Abdominal perfusion pressure \quad D. Central venous pressure }
\\
\multicolumn{6}{p{0.96\textwidth}}{\small \textbf{Answer 5}: B. Mesenteric vascular resistance }
\\
\multicolumn{6}{p{0.96\textwidth}}{\small \textbf{Question 6}: For a patient with postural abdominal pain, which sign is key to distinguishing myogenic pain? 
A. Aggravation by Valsalva maneuver \quad B. Positive rebound tenderness \quad C. Induced by postural change \quad D. Local muscle group tenderness }
\\
\multicolumn{6}{p{0.96\textwidth}}{\small \textbf{Answer 6}: D. Local muscle group tenderness }
\\
\multicolumn{6}{p{0.96\textwidth}}{\small \textbf{...}}\\
\toprule
\midrule
\multicolumn{6}{p{0.96\textwidth}}{\textbf{Diffusion with one seed data (essay-question, n = 10)}} \\
\midrule
\multicolumn{6}{p{0.96\textwidth}}{\textbf{Knowledge point path}} \\
\midrule
\multicolumn{6}{p{0.96\textwidth}}{Detection and dynamics of mini-moons around Earth}
\\
\midrule
\multicolumn{6}{p{0.96\textwidth}}{\textbf{Seed Data}}\\
\midrule
\multicolumn{6}{p{0.96\textwidth}}{\small \textbf{Question}: Is there a large number of mini-moons around Earth? Due to being too small and too dim, and at distances of over 100,000 km from Earth, could many mini-moons remain undiscovered?}
\\
\multicolumn{6}{p{0.96\textwidth}}{\small \textbf{Answer}: Currently, there is no conclusive evidence for a large number of natural mini-moons around Earth. While small objects like meter-scale asteroids can be temporarily captured by Earth's gravity (such as 2006 RH120, a 2-3 meter asteroid), these are typically transient and short-lived. No stable, naturally formed mini-moons exist in confirmed long-term orbits. The detection challenge you mention is valid - small, dim objects at great distances are difficult to observe. However, most captured objects follow unstable trajectories and eventually escape Earth's gravitational influence. As observation technology advances, we may discover more temporary companions, but a large population of stable natural mini-moons remains unlikely based on current orbital dynamics understanding.}
\\
\midrule
\multicolumn{6}{p{0.96\textwidth}}{\textbf{Synthetic QA pairs}} \\
\midrule
\multicolumn{6}{p{0.96\textwidth}}{\small \textbf{Question 1}: If a ground-based telescope has a limiting magnitude of +25, and a spherical object with albedo 0.1 is at a distance of 150,000 km from Earth, what is the minimum detectable diameter using the law of reflection (ignore atmospheric extinction, Sun-Earth distance 1 AU, visible wavelength $\lambda=550$ nm)?}
\\
\multicolumn{6}{p{0.96\textwidth}}{\small \textbf{Answer 1}: The minimum detectable diameter is about 2.4 centimeters.}
\\
\multicolumn{6}{p{0.96\textwidth}}{\small \textbf{Question 2}: Use dimensional analysis to derive the timescale $\tau$ for transient capture of mini-bodies as a function of Sun-Earth-Moon system parameters, including gravitational constant $G$, Earth's mass $M$, and lunar orbital radius $a_m$.}
\\
\multicolumn{6}{p{0.96\textwidth}}{\small \textbf{Answer 2}: $\tau \propto \sqrt{a_m^3/(GM)}$, typically weeks to months.}
\\
\multicolumn{6}{p{0.96\textwidth}}{\small \textbf{Question 3}: For a tidally-locked binary asteroid whose components are treated as homogeneous fluid ellipsoids, let the primary mass be \(M\), the secondary mass be \(m\), the orbital eccentricity be \(e\), and the mass parameter be \(\mu = m/(M+m)\).  Using the Darwin–Radau equilibrium theory, derive the approximate expression for the maximum tidal (equatorial) elongation parameter \(\delta\) in terms of \(e\) and \(\mu\).}
\\
\multicolumn{6}{p{0.96\textwidth}}{\small \textbf{Answer 3}: The maximum tidal elongation parameter is \(\displaystyle \delta \approx \frac{15}{4}\frac{\mu}{1+\mu}\,e\).}
\\
\multicolumn{6}{p{0.96\textwidth}}{\small \textbf{Question 4}: A single two-way laser-ranging station in low-Earth orbit (altitude negligible compared to target distance) is used to detect 10 cm mini-moons at a geocentric range of 1 000 km.  What is the required laser-pulse width (FWHM) for 10 cm single-shot range precision?  Ignore angular resolution and assume only time-of-flight accuracy matters.}
\\
\multicolumn{6}{p{0.96\textwidth}}{\small \textbf{Answer 4}: Required pulse width \(\tau \le 0.67\) ns.}
\\
\multicolumn{6}{p{0.96\textwidth}}{\small \textbf{Question 5}: Compare the advantages of Delaunay variables and Poincaré variables in perturbation analysis of temporary satellite orbits, and state whether their Jacobian matrices are the identity.}
\\
\multicolumn{6}{p{0.96\textwidth}}{\small \textbf{Answer 5}: Delaunay variables are convenient for Hamiltonian expansions but singular at \(e=0\) or \(i=0\); Poincaré variables remove both singularities via canonical transformation. Neither Jacobian is the identity matrix; both are symplectic.}
\\
\multicolumn{6}{p{0.96\textwidth}}{\small \textbf{Question 6}: A spherical mini-moon of radius \(R\) and bulk density \(\rho\) orbits the Sun on a circular path of radius \(a\).  Derive the secular rate of change of its semi-major axis produced by direct solar radiation pressure of strength \(P_\odot = L_\odot/(4\pi a^2 c)\), including the Bond albedo \(\beta\).}
\\
\multicolumn{6}{p{0.96\textwidth}}{\small \textbf{Answer 6}: \( \displaystyle \frac{da}{dt} = -\frac{3(1+\beta)P_\odot}{2\rho R n} \).}
\\
\multicolumn{6}{p{0.96\textwidth}}{\small \textbf{...}}\\
\toprule
\midrule
\multicolumn{6}{p{0.96\textwidth}}{\textbf{Diffusion with two seed data ( multiple-choice, n = 15)}} \\
\midrule
\multicolumn{6}{p{0.96\textwidth}}{\textbf{Knowledge point path}} \\
\midrule
\multicolumn{6}{p{0.96\textwidth}}{Construction of contrapositive statements $\rightarrow$ De Morgan's laws}
\\
\midrule
\multicolumn{3}{p{0.48\textwidth}|}{\textbf{Seed Data 1}}
 & \multicolumn{3}{|p{0.48\textwidth}}{\textbf{Seed Data 2}}\\
\midrule
\multicolumn{3}{p{0.48\textwidth}|}{\textbf{Question 1:} Which of the following statements are correct? 
A. If $p$ is a sufficient but not necessary condition for $q$, then $\neg p$ is a necessary but not sufficient condition for $\neg q$. 
B. Let $x, y \in \mathbb{R}$. The negation of the statement "If $xy=0$, then $x^2 + y^2 = 0$" is true. 
C. The contrapositive of the statement "If $xy=0$, then $x=0$ and $y=0$" is "If $x \neq 0$ and $y \neq 0$, then $xy \neq 0$". 
D. "$m = 1$ or $m = 2$" is the necessary and sufficient condition for the lines $(m+2)x + 3my + 1 = 0$ and $(m-2)x + (m+2)y - 3 = 0$ to be perpendicular. }
&
\multicolumn{3}{p{0.48\textwidth}}{\textbf{Question 2:} What is the contrapositive of a statement involving an OR condition? For example, what is the contrapositive of the statement: "For all dogs $A$, $B$, and $C$ I have, if $A$ and $B$ are male, then $B$ or $C$ are Shibas"? }
\\
\multicolumn{3}{p{0.48\textwidth}|}{\textbf{Answer 1:} The correct statements are B and C.}
&
\multicolumn{3}{p{0.48\textwidth}}{\textbf{Answer 2:} The contrapositive of "If $P$, then $Q$" is "If $\neg Q$, then $\neg P$". Use De Morgan's law: $\neg(X \text{ or } Y) = (\neg X) \text{ and } (\neg Y)$. }
\\
\midrule
\multicolumn{6}{p{0.96\textwidth}}{\textbf{Synthetic QA pairs}} \\
\midrule
\multicolumn{6}{p{0.96\textwidth}}{\small \textbf{Question 1}: What is the contrapositive of the statement 'If a number is prime greater than 2, then it is odd'? A. If a number is not odd, then it is not prime greater than 2. B. If a number is not prime greater than 2, then it is not odd. C. If a number is odd, then it is prime greater than 2. D. If a number is prime greater than 2, then it is not even. }
\\
\multicolumn{6}{p{0.96\textwidth}}{\small \textbf{Answer 1}: A. If a number is not odd, then it is not prime greater than 2. }
\\
\multicolumn{6}{p{0.96\textwidth}}{\small \textbf{Question 2}: According to De Morgan's laws, which expression is equivalent to $\neg(A \vee B)$? A. $A \wedge B$ B. $\neg A \vee \neg B$ C. $\neg A \wedge \neg B$ D. $A \vee B$ }
\\
\multicolumn{6}{p{0.96\textwidth}}{\small \textbf{Answer 2}: C. $\neg A \wedge \neg B$ }
\\
\multicolumn{6}{p{0.96\textwidth}}{\small \textbf{Question 3}: For the quantified statement '$\forall x$ (Student$(x) \rightarrow$ Studies$(x))$', what is the contrapositive? A. $\forall x (\neg \text{Student}(x) \rightarrow \neg \text{Studies}(x))$ B. $\forall x (\neg \text{Studies}(x) \rightarrow \neg \text{Student}(x))$ C. $\forall x (\text{Studies}(x) \rightarrow \text{Student}(x))$ D. $\forall x (\text{Student}(x) \wedge \neg \text{Studies}(x))$ }
\\
\multicolumn{6}{p{0.96\textwidth}}{\small \textbf{Answer 3}: B. $\forall x (\neg \text{Studies}(x) \rightarrow \neg \text{Student}(x))$ }
\\
\multicolumn{6}{p{0.96\textwidth}}{\small \textbf{Question 4}: If 'A or B' implies C, what is the contrapositive? A. If not C, then not A and not B. B. If not C, then not A or not B. C. If C, then A or B. D. If not A and not B, then not C. }
\\
\multicolumn{6}{p{0.96\textwidth}}{\small \textbf{Answer 4}:A. If not C, then not A and not B }
\\
\multicolumn{6}{p{0.96\textwidth}}{\small \textbf{Question 5}: Using De Morgan's laws, $\neg(P \wedge Q \wedge R)$ is equivalent to: A. $P \wedge Q \wedge R$ B. $\neg P \wedge \neg Q \wedge \neg R$ C. $P \vee Q \vee R$ D. $\neg P \vee \neg Q \vee \neg R$ }
\\
\multicolumn{6}{p{0.96\textwidth}}{\small \textbf{Answer 5}:D. $\neg P \vee \neg Q \vee \neg R$ }
\\
\multicolumn{6}{p{0.96\textwidth}}{\small \textbf{Question 6}: The contrapositive of 'If it is sunny or it is warm, I go outside' is: A. If I do not go outside, then it is not sunny and not warm. B. If I do not go outside, then it is not sunny or not warm. C. If I go outside, then it is sunny or warm. D. If it is not sunny and not warm, then I do not go outside. }
\\
\multicolumn{6}{p{0.96\textwidth}}{\small \textbf{Answer 6}: A. If I do not go outside, then it is not sunny and not warm. }
\\
\multicolumn{6}{p{0.96\textwidth}}{\small \textbf{...}}\\
\toprule
\midrule
\multicolumn{6}{p{0.96\textwidth}}{\textbf{Diffusion with two seed data  ( essay-question, n = 15)}} \\
\midrule
\multicolumn{6}{p{0.96\textwidth}}{\textbf{Knowledge point path}} \\
\midrule
\multicolumn{6}{p{0.96\textwidth}}{Procedures for submitting interim proposals $\rightarrow$ Rules for board meeting resolutions}
\\
\midrule
\multicolumn{3}{p{0.48\textwidth}|}{\textbf{Seed Data 1}}
 & \multicolumn{3}{p{0.48\textwidth}}{\textbf{Seed Data 2}}\\
\midrule
\multicolumn{3}{p{0.48\textwidth}|}{\textbf{Question 1:}Which of the following statements about the organizational structure of a joint stock company is incorrect? 
A. When the company's unmade-up losses reach one third of the total paid-in capital, an interim shareholders' meeting shall be convened within two months. 
B. For a joint stock company to convene a shareholders' meeting, the time, place, and matters to be considered at the meeting shall be notified to all shareholders 20 days before the meeting. 
C. Shareholders who individually or collectively hold more than 3\% of the company's shares may submit interim proposals in writing to the board of directors 10 days before the shareholders' meeting. 
D. A board meeting can only be held with more than half of the directors present, and resolutions must be passed by more than half of the attending directors.}
&
\multicolumn{3}{p{0.48\textwidth}}{\textbf{Question 2}: What are the requirements for holding a valid board meeting and passing a resolution in a joint stock company? A board meeting can only be held with more than half of the directors present, and resolutions must be passed by more than half of all directors.}
\\
\multicolumn{3}{p{0.48\textwidth}|}{\textbf{Answer 1}: A. According to corporate law, when a company's unmade-up losses reach one third of the total paid-in capital, an interim shareholders' meeting shall be convened within two months.}
&
\multicolumn{3}{p{0.48\textwidth}}{\textbf{Answer 2}: The statement is partially incorrect. A board meeting requires more than half of the directors present (correct), but resolutions must be passed by more than half of the **attending directors**, not more than half of all directors.}
\\
\midrule
\multicolumn{6}{p{0.96\textwidth}}{\textbf{Synthetic QA pairs}} \\
\midrule
\multicolumn{6}{p{0.96\textwidth}}{\small \textbf{Question 1}: What is the minimum number of supervisors required in the board of supervisors of a joint stock company?}
\\
\multicolumn{6}{p{0.96\textwidth}}{\small \textbf{Answer 1}: 3 supervisors}
\\
\multicolumn{6}{p{0.96\textwidth}}{\small \textbf{Question 2}: What is the minimum percentage of shares that a single or collective shareholder of a joint stock company must hold to request the convening of an interim shareholders' meeting?}
\\
\multicolumn{6}{p{0.96\textwidth}}{\small \textbf{Answer 2}: 3\%.}
\\
\multicolumn{6}{p{0.96\textwidth}}{\small \textbf{Question 3}: In a joint stock company, how many directors at least must be present for a board meeting to be held?}
\\
\multicolumn{6}{p{0.96\textwidth}}{\small \textbf{Answer 3}: More than half of the directors}
\\
\multicolumn{6}{p{0.96\textwidth}}{\small \textbf{Question 4}: What is the proportion of votes required for a general resolution to be passed at a shareholders' meeting in a joint stock company?}
\\
\multicolumn{6}{p{0.96\textwidth}}{\small \textbf{Answer 4}: More than half of the voting rights of shareholders present}
\\
\multicolumn{6}{p{0.96\textwidth}}{\small \textbf{Question 5}: When the company's unmade-up losses reach one third of the total paid-in capital, within how many days must an interim shareholders' meeting be convened?}
\\
\multicolumn{6}{p{0.96\textwidth}}{\small \textbf{Answer 5}: Within 60 days.}
\\
\multicolumn{6}{p{0.96\textwidth}}{\small \textbf{Question 6}: In a board meeting of a joint stock company with 11 directors and 7 present, how many directors must at least agree to pass a resolution?}
\\
\multicolumn{6}{p{0.96\textwidth}}{\small \textbf{Answer 6}: 6 directors.}
\\
\multicolumn{6}{p{0.96\textwidth}}{\small \textbf{...}}\\
\toprule
\midrule
\multicolumn{6}{p{0.96\textwidth}}{\textbf{Diffusion with three seed data (multiple-choice, n = 20)}} \\
\midrule
\multicolumn{6}{p{0.96\textwidth}}{\textbf{Knowledge point path}} \\
\midrule
\multicolumn{6}{p{0.96\textwidth}}{Poetry emotion comprehension $\rightarrow$ Imagery analysis $\rightarrow$ Theme analysis}
\\
\midrule
\multicolumn{2}{p{0.32\textwidth}|}{\textbf{Seed Data 1}} & \multicolumn{2}{p{0.32\textwidth}|}{\textbf{Seed Data 2}} & \multicolumn{2}{p{0.32\textwidth}}{\textbf{Seed Data 3}}\\
\midrule
\multicolumn{2}{p{0.32\textwidth}|}{\small \textbf{Question 1}: Read the following Song dynasty poem and answer: Which of the following appreciations are NOT appropriate? A. The first line uses 'crown and robe' to refer to the emperor, who is surrounded by ministers and stands above the people. B. The word 'only' in the second line depicts the ministers' cowardice. C. The couplet only praises Yongshu for his fairness, but also criticizes his lack of reason and rashness. D. The ending expresses the poet's best wishes for Yongshu. E. The poem is precisely constructed, deep in language, integrating narration, emotion, description, and commentary.}
& \multicolumn{2}{p{0.32\textwidth}|}{\small \textbf{Question 2}: Read the following Tang poem 'Wind and Rain' by Li Shangyin. What is the main emotion expressed in the poem? }
& \multicolumn{2}{p{0.32\textwidth}}{\small \textbf{Question 3}: Read the following poem 'Apricot Blossoms by the North Slope' by Wang Anshi. What is the key imagery and philosophical meaning in the line 'Even if blown into snow by the spring wind'?}
\\
\multicolumn{2}{p{0.32\textwidth}|}{\small \textbf{Answer 1}: C, D}
& \multicolumn{2}{p{0.32\textwidth}|}{\small \textbf{Answer 2}: Melancholy, loneliness, and the hardships of life in exile.}
& \multicolumn{2}{p{0.32\textwidth}}{\small \textbf{Answer 3}: The image of apricot blossoms as snow reflects a realm of subjective struggle and philosophical transcendence.}
\\
\midrule
\multicolumn{6}{p{0.96\textwidth}}{\textbf{Synthetic QA pairs}} \\
\midrule
\multicolumn{6}{p{0.96\textwidth}}{\small \textbf{Question 1}: When comparing 'Guo po shan he zai' from Du Fu's 'Spring View' and 'Huang ye reng feng yu' from Li Shangyin's 'Wind and Rain', what is the common core technique and emotional effect? A. Both use virtual-real combination to express loneliness B. Both use contrast to highlight social injustice C. Both use natural imagery to symbolize national decline and strengthen patriotic feelings D. Both use allusions to express frustration at unrecognized talent}
\\
\multicolumn{6}{p{0.96\textwidth}}{\small \textbf{Answer 1}: C. Both use natural imagery to symbolize national decline and strengthen patriotic feelings.}
\\
\multicolumn{6}{p{0.96\textwidth}}{\small \textbf{Question 2}: According to Wang Guowei's 'theory of realms', which realm is best reflected in Wang Anshi's 'Even if blown into snow by the spring wind' in 'Apricot Blossoms by the North Slope'? A. Realm of no-self, emphasizing the unity of subject and object B. Realm of self, highlighting subjective struggle C. Writing the scene, focusing on realistic depiction D. Creating a realm, constructing an ideal world}
\\
\multicolumn{6}{p{0.96\textwidth}}{\small \textbf{Answer 2}: A. Realm of no-self, emphasizing the unity of subject and object.}
\\
\multicolumn{6}{p{0.96\textwidth}}{\small \textbf{Question 3}: The fundamental reason for the difference in opening techniques between Su Shi's 'The River Flows East' in 'Reminiscing at Red Cliff' and Xin Qiji's 'Eternal Joy at Jingkou North Pavilion' 'Thousand-year-old rivers and mountains' is? A. Su Shi emphasizes the passage of time, Xin Qiji emphasizes vastness of space B. Su Shi uses virtual description, Xin Qiji uses realistic description C. Su Shi expresses personal detachment, Xin Qiji emphasizes historical responsibility D. Su Shi is influenced by Taoism, Xin Qiji by Confucianism}
\\
\multicolumn{6}{p{0.96\textwidth}}{\small \textbf{Answer 3}: C. Su Shi expresses personal detachment, Xin Qiji emphasizes historical responsibility}
\\
\multicolumn{6}{p{0.96\textwidth}}{\small \textbf{Question 4}: Both Li Bai's 'Born to be useful' in 'Bring in the Wine' and Li Shangyin's 'Heartbroken at the wine of Xinfeng' in 'Wind and Rain' mention wine, but the difference in their attitudes towards life lies in? A. Li Bai is optimistic and enterprising, Li Shangyin is pessimistic and depressed B. Li Bai writes about the grandeur of the High Tang, Li Shangyin about the decline of the Late Tang C. Li Bai uses wine to express heroism, Li Shangyin uses wine to express sorrow D. Li Bai focuses on personal enjoyment, Li Shangyin focuses on social criticism}
\\
\multicolumn{6}{p{0.96\textwidth}}{\small \textbf{Answer 4}: C. Li Bai uses wine to express heroism, Li Shangyin uses wine to express sorrow}
\\
\multicolumn{6}{p{0.96\textwidth}}{\small \textbf{Question 5}: The artistic conception created by Wang Wei's 'After new rain in the empty mountains' in 'Autumn Evening in the Mountains' is similar to Tao Yuanming's 'Picking chrysanthemums by the eastern fence' in 'Drinking Wine', but the key philosophical difference is? A. Wang Wei blends Zen emptiness, Tao Yuanming embodies Confucian engagement B. Wang Wei emphasizes natural beauty, Tao Yuanming highlights the joy of seclusion C. Wang Wei is influenced by Buddhism, Tao Yuanming by Daoist non-action D. Wang Wei writes realistically, Tao Yuanming writes idealistically}
\\
\multicolumn{6}{p{0.96\textwidth}}{\small \textbf{Answer 5}: C. Wang Wei is influenced by Buddhism, Tao Yuanming by Daoist non-action}
\\
\multicolumn{6}{p{0.96\textwidth}}{\small \textbf{Question 6}: In Bai Juyi's 'The Song of the Lute Player', how is the emotional effect of 'We are all people fallen to the ends of the earth' mainly enhanced? A. By contrasting the fates of the musician and the poet B. By simulating life's ups and downs through musical rhythm C. By using the imagery of the ends of the earth to symbolize a wandering fate D. By direct expression to strengthen resonance}
\\
\multicolumn{6}{p{0.96\textwidth}}{\small \textbf{Answer 6}: A. By contrasting the fates of the musician and the poet}
\\
\multicolumn{6}{p{0.96\textwidth}}{\small \textbf{...}}\\
\toprule
\midrule
\multicolumn{6}{p{0.96\textwidth}}{\textbf{Diffusion with three seed data (essay-question, n = 20)}} \\
\midrule
\multicolumn{6}{p{0.96\textwidth}}{\textbf{Knowledge point path}} \\
\midrule
\multicolumn{6}{p{0.96\textwidth}}{Manifestations of racial segregation policy in the military $\rightarrow$ Colonial troops in World War I $\rightarrow$ Wartime social conditions and soldier adaptation}
\\
\midrule
\multicolumn{2}{p{0.32\textwidth}|}{\textbf{Seed Data 1}} & \multicolumn{2}{p{0.32\textwidth}|}{\textbf{Seed Data 2}} & \multicolumn{2}{p{0.32\textwidth}}{\textbf{Seed Data 3}}\\
\midrule
\multicolumn{2}{p{0.32\textwidth}|}{\small \textbf{Question 1}: What organization was Doris "Dorie" Miller a prominent figure in and what was his significant action during a critical event in that organization's history? }
& \multicolumn{2}{p{0.32\textwidth}|}{\small \textbf{Question 2}: Just another relapse post. (Long narrative about PTSD, trauma, and relapse during isolation, with emphasis on lack of support and coping mechanisms.)}
& \multicolumn{2}{p{0.32\textwidth}}{\small \textbf{Question 3}: What role did colonial troops play in World War I and how did wartime social conditions affect soldier adaptation?}
\\
\multicolumn{2}{p{0.32\textwidth}|}{\small \textbf{Answer 1}: Doris "Dorie" Miller was a prominent figure in the United States Navy during World War II. He is best known for his heroic actions during the attack on Pearl Harbor, where he manned an anti-aircraft gun without prior training and assisted wounded sailors, becoming a symbol of courage and challenging racial stereotypes.}
& \multicolumn{2}{p{0.32\textwidth}|}{\small \textbf{Answer 2}: The narrative details a PTSD relapse triggered by isolation, referencing trauma, institutional abuse, and the importance of support systems and coping strategies.}
& \multicolumn{2}{p{0.32\textwidth}}{\small \textbf{Answer 3}: Colonial troops played significant roles in World War I, often facing racial discrimination and challenging social conditions, which impacted their adaptation and psychological well-being.}
\\
\midrule
\multicolumn{6}{p{0.96\textwidth}}{\textbf{Synthetic QA pairs}} \\
\midrule
\multicolumn{6}{p{0.96\textwidth}}{\small \textbf{Question 1}: A study on PTSD relapse rates during COVID-19 isolation found that 120 out of 200 participants with prior trauma history relapsed into self-harm. Calculate the odds ratio for relapse if 70 out of 150 participants without isolation triggers relapsed. Interpret the clinical significance.}
\\
\multicolumn{6}{p{0.96\textwidth}}{\small \textbf{Answer 1}: Odds ratio = 1.714; Isolation triggers significantly increase relapse risk.}
\\
\multicolumn{6}{p{0.96\textwidth}}{\small \textbf{Question 2}: Using DSM-5 criteria, diagnose PTSD for a veteran who experiences avoidance of war reminders, hypervigilance, and negative alterations in cognition for 3 months after a combat event. One symptom from each cluster must be specified.}
\\
\multicolumn{6}{p{0.96\textwidth}}{\small \textbf{Answer 2}: No, PTSD diagnosis not met due to missing intrusion symptoms.}
\\
\multicolumn{6}{p{0.96\textwidth}}{\small \textbf{Question 3}: In a historical analysis, if 30\% of African American sailors during WWII were restricted to menial roles due to segregation policies, and Doris Miller's action increased public support for desegregation by 40\% based on media archives, calculate the probability that a random policy change within 5 years was influenced by Miller. Assume base desegregation support was 20\%.}
\\
\multicolumn{6}{p{0.96\textwidth}}{\small \textbf{Answer 3}: Estimated probability = 40\%.}
\\
\multicolumn{6}{p{0.96\textwidth}}{\small \textbf{Question 4}: A cognitive-behavioral therapy (CBT) trial for PTSD had a 75\% success rate in a sample of 80 patients. Calculate the 95\% confidence interval for the true success rate and determine if it is statistically different from a control group rate of 60\% (n=80). Use z-test for proportions.}
\\
\multicolumn{6}{p{0.96\textwidth}}{\small \textbf{Answer 4}: 95\% CI = (0.655, 0.845); Yes, significantly different from control.}
\\
\multicolumn{6}{p{0.96\textwidth}}{\small \textbf{Question 5}: Based on the UN Convention Against Torture (CAT) definition—‘severe pain or suffering, physical or mental, intentionally inflicted by or with the consent of a public official for purposes such as punishment, coercion or discrimination’—which scenario meets the criteria: (a) Solitary confinement in a hospital for 48 hours with no medical justification and under color of law, or (b) Verbal abuse by a family member?}
\\
\multicolumn{6}{p{0.96\textwidth}}{\small \textbf{Answer 5}: Scenario (a) meets the UN torture criteria; (b) lacks state involvement and severity required under CAT.}
\\
\multicolumn{6}{p{0.96\textwidth}}{\small \textbf{Question 6}: If the relapse rate for self-harm is 40\% in PTSD patients without intervention, and CBT reduces it by 25\% relative risk, calculate the absolute risk reduction (ARR) and number needed to treat (NNT) for CBT.}
\\
\multicolumn{6}{p{0.96\textwidth}}{\small \textbf{Answer 6}: ARR = 10\%; NNT = 10.}\\
\multicolumn{6}{p{0.96\textwidth}}{\small \textbf{...}}\\

\end{longtable}

\begin{longtable}{p{0.98\textwidth}}
\caption{Case of different difficulty of generate data.}
\label{tab:case_hlabel_syn} \\
\toprule
\midrule
\endfirsthead

\caption[]{(continued)} \\
\toprule
\midrule
\endhead

\midrule
\endfoot

\bottomrule
\endlastfoot

\small \textbf{H1} \\
\midrule
\small \textit{\textbf{Sample 1}:} Question: An antibacterial drug achieves therapeutic target value when AUC/MIC=125. Given MIC=2mg/L, CL=10L/h, what is the total daily dose (F=1)? \\
\small Answer: Total daily dose is 2500mg \\
\small \textit{\textbf{Sample 2}:} Question: When merging two sorted singly linked lists into one sorted list, what is the minimum number of pointer reassignments required in the worst-case scenario? \\
\small A. O(1)    B. O(log n)    C. O(n)    D. O(n log n) \\
\small Answer: C. O(n) \\
\midrule

\small \textbf{H2} \\
\midrule
\small \textit{\textbf{Sample 1}:} Question: A company's current stock price is \$80, with expected earnings per share of \$5.00 next year and a dividend payout ratio of 40\%. If the market required rate of return is 10\%, calculate the implied perpetual growth rate. If the actual growth rate is 5\%, determine if the stock price is reasonable. \\
\small Answer: The implied growth rate is 7.5\%, and the current stock price is overvalued. \\
\small \textit{\textbf{Sample 2}:} Question: If using game theory to model the collapse of the feudal system, which equilibrium best describes the struggle for power among lords? \\
\small A. Nash equilibrium (individual optimality leading to collective suboptimality)    B. Pareto optimality (no improvement without harming others)    C. Coordination game (cooperation driven by common interests)    D. Prisoner's dilemma (betrayal as dominant strategy) \\
\small Answer: D. Prisoner's dilemma (betrayal as dominant strategy) \\
\midrule

\small \textbf{H3} \\
\midrule
\small \textit{\textbf{Sample 1}:} Question: Design a machine learning-based algorithm to predict skin retraction rate after liposuction. Which biomechanical parameters should be selected? Explain the parameter selection criteria and algorithm architecture. \\
\small Answer: Key parameters include biomechanical factors (initial dermal strain $\varepsilon_0$, stress relaxation time $\tau$, elastic modulus $E$, skin thickness), physiological factors (age, BMI, collagen density, skin hydration), and surgical parameters (liposuction volume, anatomical location). Parameter selection criteria: clinical relevance, biomechanical significance, and measurability. Architecture: data preprocessing → feature engineering → ensemble model (XGBoost + Random Forest) → output layer (retraction rate \%) → model interpretation layer. The model incorporates 15+ parameters with cross-validation and SHAP analysis for clinical interpretability. \\
\small \textit{\textbf{Sample 2}:} Question: Given two coprime integers m and n, what is the minimal k such that there exist unique pairs (x,y) in \{1,2,...,k\} satisfying x $\equiv$ a mod m and y $\equiv$ b mod n? \\
\small A. m + n    B. m + n - 1    C. m * n    D. lcm(m,n) \\
\small Answer: B. m + n - 1 \\
\midrule

\small \textbf{H4} \\
\midrule
\small \textit{\textbf{Sample 1}:} Question: A child with congenital hypothyroidism still shows delayed intellectual development after treatment. Analyze possible reasons from a molecular mechanism perspective. \\
\small Answer: Possible mechanisms: 1: MCT8 transporter defects causing T3 deficiency in brain tissue; 2: THRB mutations causing hormone resistance; 3: Treatment initiated after the critical period of 2 weeks after birth. \\
\small \textit{\textbf{Sample 2}:} Question: Matrix A = [[1, 2], [0, 1]] transforms the curve x$^2$ + y$^2$ = 1 into a new curve C. What is the minimum distance between points on C and the line x - y = 3? \\
\small A. (3 - $\sqrt{2}$)/$\sqrt{2}$    B. $\sqrt{2}$    C. 3/$\sqrt{2}$    D. 3 - $\sqrt{2}$ \\
\small Answer: A. (3 - $\sqrt{2}$)/$\sqrt{2}$ \\
\midrule

\small \textbf{H5} \\
\midrule
\small \textit{\textbf{Sample 1}:} Question: Formulate the hydrodynamic limit of the Boltzmann equation to the incompressible Navier-Stokes equations using the Chapman-Enskog expansion. Derive the viscosity and heat conductivity coefficients in terms of the collision kernel. \\
\small Answer: Viscosity $\mu = \frac{1}{15} \int_{\mathbb{R}^3} \int_{S^2} |v - v_*|^\gamma b(\cos\theta) \sin^2\theta d\theta dv$ and heat conductivity $\kappa = \frac{5}{2}\mu$ for Maxwell molecules ($\gamma=1$), following the Chapman-Enskog procedure. \\
\small \textit{\textbf{Sample 2}:} Question: When constructing the contraction $\phi_{|A|}: \overline{M}_{0,n}(X,\beta) \to Y_{|A|}$, if $A$ is a Kawamata divisor class on $\overline{M}_{0,n+m}/\mathfrak{S}_m$, what is the necessary condition for the Picard rank of $Y_{|A|}$ to be 1? \\
\small A. $\beta$ is an extremal curve class    B. Picard rank of $X$ is 1    C. $\mathfrak{S}_m$ acts freely    D. All boundary divisors $\Delta_{i,j}$ are contracted ($i,j\neq 2,n+m-2$) \\
\small Answer: D. All boundary divisors $\Delta_{i,j}$ are contracted ($i,j\neq 2,n+m-2$) \\

\end{longtable}

\begin{longtable}{p{0.98\textwidth}}
\caption{Samples for synthetic QA pairs of different disciplines. For each discipline, Sample 1 is multiple-choice and Sample 2 is essay- question.}
\label{tab:case_discipline} \\
\toprule
\midrule
\endfirsthead

\caption[]{(continued)} \\
\toprule
\midrule
\endhead

\midrule
\endfoot

\bottomrule
\endlastfoot

\small \textbf{Mathematics} \\
\midrule
\small \textit{\textbf{Sample 1}:} Question: An e-commerce platform has discount rules: 50 yuan off when spending 299 yuan, 100 yuan off when spending 499 yuan, with coupons stackable. For an order with items totaling 630 yuan and shipping fee of 15 yuan, what is the minimum payment amount after applying discounts? \\
\small Answer: 545 \\
\small \textit{\textbf{Sample 2}:} Question: Which principle ensures that the outer measure of the set of random reals is 1 in both V and V[G], preventing it from being 0? \\
\small A. Baire Category Theorem    B. Kolmogorov's Zero-One Law    C. Fubini's Theorem    D. Borel Determinacy \\
\small Answer: A. Baire Category Theorem \\
\midrule

\small \textbf{Computer Science and Technology} \\
\midrule
\small \textit{\textbf{Sample 1}:} Question: In 3D stacked chips, analyze the impact of Through-Silicon Via (TSV) on clock signal integrity, and propose three methods to suppress reflection noise with their comparative advantages and disadvantages. \\
\small Answer: Termination resistors (stable but area-consuming), stepped TSV (efficient but difficult to fabricate), pre-emphasis (flexible but requires additional circuitry). \\
\small \textit{\textbf{Sample 2}:} Question: How to solve the spatial audio mismatch problem caused by audio-video separation when processing VR 360° videos? \\
\small A. Add Ambisonic metadata    B. Force mono audio output    C. Enable head-related transfer function    D. Increase spatial audio delay compensation \\
\small Answer: A. Add Ambisonic metadata \\
\midrule

\small \textbf{Clinical Medicine} \\
\midrule
\small \textit{\textbf{Sample 1}:} Question: For a newborn with bilateral persistent eye discharge without redness or fever, how should anatomical characteristics of the lacrimal duct and microbiological features be integrated for differential diagnosis? \\
\small Answer: Analysis must consider the unopened Hasner's valve at the distal nasolacrimal duct (causing sterile mucus accumulation) and bacterial colonization risks (such as C. trachomatis vertical infection incubation characteristics), using cytological examination of secretions (neutrophil predominance suggests infection) and pathogen culture for differentiation. \\
\small \textit{\textbf{Sample 2}:} Question: When comparing the response differences to LASIK surgery between children and adult amblyopia patients, which of the following is the most fundamental reason for poorer prognosis in adults? \\
\small A. Decreased corneal healing capacity    B. Closure of neural plasticity window    C. Lower refractive error stability    D. Reduced visual system redundancy \\
\small Answer: B. Closure of neural plasticity window \\
\midrule

\small \textbf{Chemistry} \\
\midrule
\small \textit{\textbf{Sample 1}:} Question: Calculate the Gibbs free energy change ($\Delta G$) during zymogen activation, given that peptide bond hydrolysis releases -5 kJ/mol and conformational reorganization requires +3 kJ/mol. Determine if the process is spontaneous and explain its physiological significance. \\
\small Answer: $\Delta G = -2$ kJ/mol $< 0$, so the process is spontaneous. Physiological significance: The hydrolysis reaction drives conformational changes, ensuring activation occurs only under specific triggering conditions, preventing accidental activation. \\
\small \textit{\textbf{Sample 2}:} Question: Given the battery reaction: Zn + 2H$^+$ $\rightarrow$ Zn$^{2+}$ + H$_2$, with standard electromotive force of 0.76 V. If conducted in a buffer solution at pH=5, with P$_{H_2}$=1 atm and [Zn$^{2+}$]=0.1 M, the actual electromotive force is approximately: (Given E$^{\circ}_{Zn^{2+}/Zn}$ = -0.76 V) \\
\small A. 0.50 V    B. 0.64 V    C. 0.76 V    D. 0.88 V \\
\small Answer: A. 0.50 V \\
\midrule

\small \textbf{Economics} \\
\midrule
\small \textit{\textbf{Sample 1}:} Question: After a country implements a universal basic income policy, how can one distinguish between the direct economic effects and indirect health awareness effects of this policy on the increased use of preventive medical services? Please propose a regression model specification that includes instrumental variables. \\
\small Answer: Use a 2SLS model with pre-implementation regional characteristics as instrumental variables, treating total income change as an endogenous variable. By comparing the significance of economic variable coefficients with health awareness variable coefficients, decompose direct economic effects and indirect awareness effects. \\
\small \textit{\textbf{Sample 2}:} Question: In an environment of rising interest rate expectations, a company plans to buy back stocks and hold floating-rate debt. What is the most effective derivative strategy to hedge the overall risk? \\
\small A. Buy Interest Rate Cap    B. Sell Interest Rate Swap (Pay Fixed, Receive Floating)    C. Buy Treasury futures    D. Sell Credit Default Swap (CDS) \\
\small Answer: A. Buy Interest Rate Cap \\
\midrule

\small \textbf{Information Science and Systems Science} \\
\midrule
\small \textit{\textbf{Sample 1}:} Question: For metal parts in additive manufacturing, how to build a digital twin system for quality prediction based on acoustic emission signals? Explain the feature extraction algorithm and real-time simulation architecture. \\
\small Answer: Key features: $1:$ Energy ratio in the 100-150kHz frequency band $2:$ Hilbert-Huang marginal spectral entropy value $3:$ Principal components from singular value decomposition of joint time-frequency distribution. \\
\small \textit{\textbf{Sample 2}:} Question: In a negative feedback system with open-loop transfer function $G(s) = \frac{K}{s(s+2)}$, where $K > 0$, what condition must $K$ satisfy for the closed-loop system to be stable? \\
\small A. $K < 2$ \quad B. $K > 2$ \quad C. $K < 4$ \quad D. $K > 4$ \\
\small Answer: D. $K > 4$ \\
\midrule

\small \textbf{Physics} \\
\midrule
\small \textit{\textbf{Sample 1}:} Question: In the relativistic framework, when an object moves at 0.8c, what is the ratio of its relativistic mass to rest mass? If a force perpendicular to the direction of motion is applied at this time, derive the expression for acceleration. \\
\small Answer: Relativistic mass ratio is 5/3; transverse acceleration a$_{\perp}$ = 3F/(5m$_0$) \\
\small \textit{\textbf{Sample 2}:} Question: Gravity is described as spacetime curvature, while electromagnetic force is transmitted through photons. What is the key contradiction when unifying these two forces at the quantum scale? Which higher-order theoretical framework needs to be introduced? \\
\small A. Contradiction in force range; string theory    B. Contradiction in relativistic effects; quantum field theory    C. Contradiction in energy conservation; grand unified theory    D. Contradiction in time asymmetry; loop quantum gravity \\
\small Answer: A. Contradiction in force range; string theory \\
\midrule

\small \textbf{Biology} \\
\midrule
\small \textit{\textbf{Sample 1}:} Question: Explain why a protein solution that has been heat-sterilized might regain activity after cooling, while a protein solution treated with pepsin cannot recover its activity. \\
\small Answer: Heat denaturation doesn't break covalent bonds, allowing potential renaturation upon cooling; pepsin digestion hydrolyzes peptide bonds causing irreversible structural damage. \\
\small \textit{\textbf{Sample 2}:} Question: Comparing the developmental cycles of Chlamydia and Rickettsia, what is the key difference in their reproductive morphology energy metabolism? Integrate biochemical pathway reasoning. \\
\small A. Chlamydial elementary bodies depend on host ATP, while Rickettsia perform autonomous oxidative phosphorylation    B. Both utilize glycolysis for energy, but Chlamydia at higher rates    C. Rickettsia reticulate bodies have mitochondria-like structures, while Chlamydia lack them    D. Chlamydia utilize photosynthesis, while Rickettsia depend on amino acid degradation \\
\small Answer: A. Chlamydial elementary bodies depend on host ATP, while Rickettsia perform autonomous oxidative phosphorylation \\
\midrule

\small \textbf{Law} \\
\midrule
\small \textit{\textbf{Sample 1}:} Question: According to relevant provisions of the Company Law and Securities Law, if Company A controls 60\% of Company B's voting rights through an agreement, and Company B holds 15\% shares of listed Company C. When Company A increases its shareholding in Company C to 8\%, is it required to make a tender offer? Please analyze based on the rules for identifying persons acting in concert. \\
\small Answer: Triggered. A and B constitute persons acting in concert, with combined shareholding of 23\% (15\%+8\%). Although below 30\%, after the increase, every additional 5\% requires suspension and announcement. The trap in this question is the cumulative calculation rule and the regulation of shareholding increases. \\
\small \textit{\textbf{Sample 2}:} Question: Limited partner Qian engaged in a competitive business and profited, but used the partnership's trade secrets. General partner Sun claimed the profits belonged to the partnership. If the partnership agreement does not stipulate this situation, according to the Anti-Unfair Competition Law and Partnership Enterprise Law, which of the following is correct? \\
\small A. All profits belong to the partnership due to infringement    B. Part of the profits belong to the partnership, with the proportion determined by court    C. Qian bears no responsibility as competitive business is implicitly allowed    D. Sun must prove intentional infringement to seek compensation \\
\small Answer: B. Part of the profits belong to the partnership, with the proportion determined by court \\

\end{longtable}

\begin{longtable}{p{0.98\textwidth}}
\caption{Samples for synthetic QA pairs of different knowledge points. For each discipline, Sample 1 is multiple-choice and Sample 2 is essay- question.}
\label{tab:case_knowledge_point} \\
\toprule
\midrule
\endfirsthead

\caption[]{(continued)} \\
\toprule
\midrule
\endhead

\midrule
\endfoot

\bottomrule
\endlastfoot

\small \textbf{Single Knowledge Point} \\
\midrule
\small \textit{\textbf{Sample 1}:}\\
\small \textit{Knowledge Point:} Legal Reservation Principle \\
\small Question: An art gallery requires teachers in group visits to present an introduction letter from their institution, while students only need student IDs. Some teachers sued for infringement of equal rights. Please analyze the legality of the introduction letter requirement based on the principle of legal reservation. \\
\small Answer: Illegal. The requirement for an introduction letter lacks legal basis and constitutes an undue restriction on teachers' rights, violating the principle of administrative legality. \\
\midrule
\small \textit{\textbf{Sample 2}:}\\
\small \textit{Knowledge Point:} Noether Normalization Lemma \\
\small Question: Noether normalization lemma states that a finitely generated k-algebra contains a polynomial subalgebra over which it is integral. In the geometric interpretation, what does this imply about the corresponding variety? \\
\small A. It is unirational    B. It is affine and irreducible    C. It has a finite morphism to affine space    D. It is smooth in codimension one \\
\small Answer: C. It has a finite morphism to affine space \\
\midrule

\small \textbf{Two Knowledge Points} \\
\midrule
\small \textit{\textbf{Sample 1}:} \\
\small \textit{Knowledge Points:} Acid-Base Balance Disorders, Water-Electrolyte Balance Disorders \\
\small Question: A patient with respiratory failure developed altered consciousness after mechanical ventilation. Emergency tests showed: blood sodium 128mmol/L, chloride 88mmol/L, arterial blood gas pH 7.52, PaCO$_2$ 28mmHg, HCO$_3^-$ 24mmol/L. Diagnose the acid-base imbalance type and explain the mechanism of ionic abnormalities. \\
\small Answer: Acute respiratory alkalosis; low sodium and chloride suggest concurrent dilutional hyponatremia, possibly due to inappropriate antidiuretic hormone secretion after ventilation improvement. \\
\midrule
\small \textit{\textbf{Sample 2}:}\\
\small \textit{Knowledge Points:} AIC Model Selection Criteria, Linear Regression Model Assumptions \\
\small Question: When comparing nested linear models (Model A: basic features, Model B: adds polynomial terms) using AIC, which outcome indicates that the polynomial terms are justified, and what is a key assumption? \\
\small A. AIC decreases by more than 2, assuming errors are normally distributed and independent.    B. AIC increases, but p-values for polynomial terms are significant, indicating overfitting.    C. R-squared improves marginally, requiring BIC to confirm model parsimony.    D. F-statistic is insignificant, implying polynomial terms add no value despite AIC change. \\
\small Answer: A. AIC decreases by more than 2, assuming errors are normally distributed and independent. \\
\midrule

\small \textbf{Three Knowledge Points} \\
\midrule
\small \textit{\textbf{Sample 1}:}\\
\small \textit{Knowledge Points:} Phase Field Method Fundamentals, Anisotropic Phase Field Models, Free Energy Function Construction \\
\small Question: Within the phase field method framework, derive the control equations for a polycrystalline growth model that generates Voronoi-like structures. With grain boundary energy anisotropy coefficient $<0.05$, provide the phase field variable evolution equation and free energy function form. \\
\small Answer: The control equation is the anisotropic Allen-Cahn equation, with a free energy function containing gradient terms, double-well potential, and intergranular repulsion terms. Anisotropy is modulated through $\varepsilon(\theta)$. \\
\midrule
\small \textit{\textbf{Sample 2}:}\\
\small \textit{Knowledge Points:} Isolating Switch Operation Risk Analysis, Power System Transient Process, Surge Arrester and Voltage Transformer Configuration Principles \\
\small Question: In a 220kV double-busbar connection, if surge arresters and voltage transformers share one set of isolating switches, the maximum operational risk occurs when: \\
\small A. The isolating switch operates under load and produces an arc    B. Voltage fluctuation caused by surge arrester operation during lightning strikes    C. Short circuit on the secondary side of voltage transformer    D. Transient process during busbar switching \\
\small Answer: A. The isolating switch operates under load and produces an arc \\

\end{longtable}

\end{document}